%% file: main.tex
\pdfoutput=1
\documentclass[11pt]{article}

\usepackage[]{acl}

\usepackage{times}
\usepackage{latexsym}

\usepackage{tcolorbox}
\usepackage{enumitem}

\usepackage{makecell}
\usepackage[T1]{fontenc}
\usepackage[utf8]{inputenc}

\usepackage{microtype}
\usepackage{inconsolata}

\usepackage{booktabs}
\newcommand{\mysubsection}[1]{\vspace{0.3em}\noindent\textbf{#1}}

\usepackage{tikz}
\usepackage{pgfplots}
\usepackage{multirow}
\pgfplotsset{compat=newest}
\usepgfplotslibrary{colorbrewer}
\usepgfplotslibrary{statistics}
\usetikzlibrary{patterns}
\newcommand{\sref}[1]{\S\ref{#1}}

\DeclareUnicodeCharacter{25B3}{$\bigtriangleup$}
\DeclareUnicodeCharacter{25BD}{$\bigtriangledown$}

\usepackage{svg}

\title{The Muddy Waters of Modeling Empathy in Language:\\The Practical Impacts of Theoretical Constructs}

\author{Allison Lahnala, Charles Welch, David Jurgens, Lucie Flek}

\author{Allison Lahnala$^\diamondsuit$ \and Charles Welch$^\heartsuit$ \and David Jurgens$^\spadesuit$ \and Lucie Flek$^\diamondsuit$ \\ 
    $^\diamondsuit$University of Bonn, $^\heartsuit$McMaster University, 
     $^\spadesuit$University of Michigan \\
    \texttt{alahnala@gmail.com},
    \texttt{cwelch@mcmaster.ca}, 
    \texttt{jurgens@umich.edu},
    \texttt{flek@bit.uni-bonn.de}, 
}

\begin{document}
\maketitle
\begin{abstract}

Conceptual operationalizations of empathy in NLP are varied, with some having specific behaviors and properties, while others are more abstract. How these variations relate to one another and capture properties of empathy observable in text remains unclear. To provide insight into this, we analyze the transfer performance of empathy models adapted to empathy tasks with different theoretical groundings. We study (1) the dimensionality of empathy definitions, (2) the correspondence between the defined dimensions and measured/observed properties, and (3) the conduciveness of the data to represent them, finding they have a significant impact to performance compared to other transfer setting features. Characterizing the theoretical grounding of empathy tasks as \textit{direct}, \textit{abstract}, or \textit{adjacent} further indicates that tasks that directly predict specified empathy components have higher transferability. Our work provides empirical evidence for the need for precise and multidimensional empathy operationalizations.
\end{abstract}

\section{Introduction}

Empathy is considered a desirable aspect of interactions, but the difficulty of defining and operationalizing it inhibits scientific inquiry into the construct~\cite{hall-schwartz-2019-empathy} and its role  (e.g., in clinical encounters~\cite{neumann_2009_clincial_empathy}). Nevertheless, empathy is increasingly desired for conversational agents~\cite{chen-etal-2023-soulchat}, leading many NLP researchers to investigate generative models of empathy, while appropriate evaluation frameworks are still limited~\cite{lee-etal-2024-comparative}. It is therefore a critical time to empirically investigate whether existing NLP operationalizations can provide a foundation for robust evaluation frameworks of empathic language, whether for examining empathy in generated output or human conversations. 

\begin{figure}
    \centering
    \includegraphics[width=\linewidth]{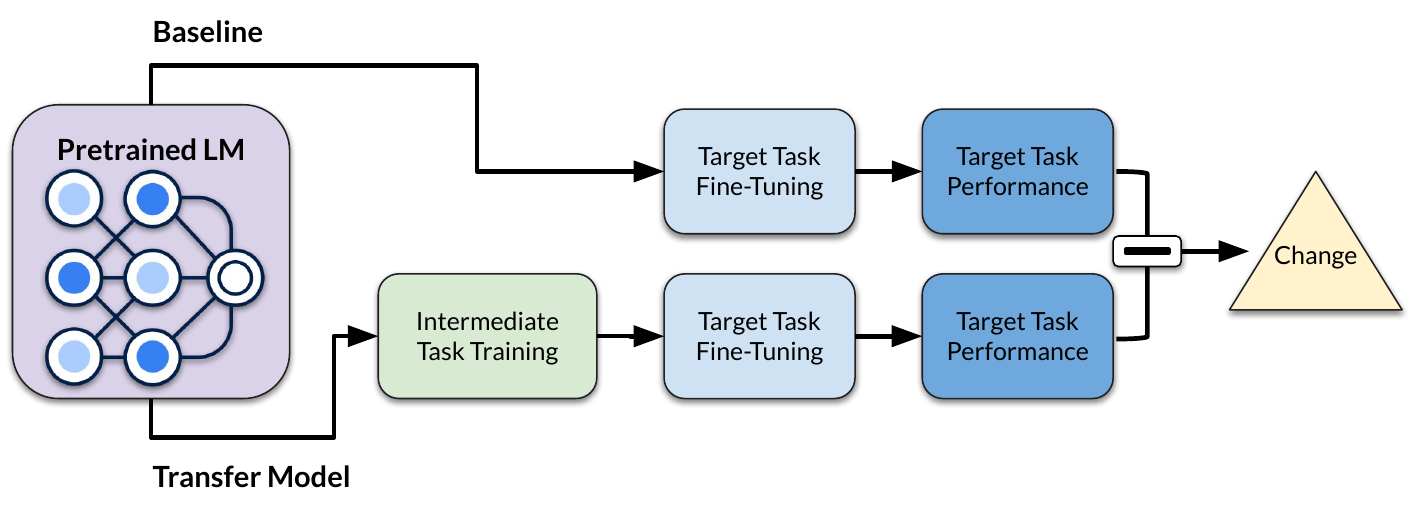}
    \caption{Experimental setup: How does intermediate task training on one empathy task impact performance on another target empathy task?}
    \label{fig:paper-fig}
\end{figure}

NLP research on empathy detection and generation generally relies on the notion that facets of empathy may be measured or observed through language, whether they are considered a trait~\cite{litvak-etal-2016-social,yaden2024characterizing}, a state ~\cite{buechel-etal-2018-modeling}, or a social function in conversations \cite{omitaomu2022empathic,zhou-jurgens-2020-condolence}. 
Empathy operationalizations in NLP range from singular ratings of empathy or emotion mirroring to fine-grained indicators and behaviors. Yet the underlying theoretical constructs and their mappings into language-based measurement tend to be underspecified, making it unclear whether our models effectively represent empathy and its components \cite{shetty2023scoping}. Given the often vague, varied use of ``empathy'' in NLP literature, it is not well understood how different operationalizations relate or what aspects of empathy they model ~\cite{lahnala-etal-2022-acritical}.

We aim to empirically quantify the practical effects of conceptual operationalizations to provide insight for developing future language-based empathy measurements and resources.
We approach this with \textit{transfer learning} experiments between 18 empathy tasks, hypothesizing that models trained on corpora with more similar constructs should provide more benefit to the target tasks~\cite{alyafeai2020survey,luo-etal-2022-cogtaskonomy}. Thus, transfer performance provides empirical indicators for answering our primary research question:

\mysubsection{How does the theoretical grounding of empathy tasks impact transfer performance?} 

\noindent First, we assess the theoretical grounding of empathy operationalizations according to their definition \textit{granularity}, \textit{correspondence} between the operationalized measurements or observations to the definition, and \textit{conduciveness} of the language situation to represent empathy in the data, along 5-point Likert scales. 
We identify three themes of the selected tasks as predicting aspects of the conceptualization \textit{directly}, an \textit{abstract} representation of empathy, or a construct \textit{adjacent} to empathy. 
Then, each task is tested as an \textit{intermediate} task for each other task as a \textit{target} task in the transfer model setup shown in Figure  \ref{fig:paper-fig}, compared to a model without intermediate training.

We examine impacts to target task performance by considering the transfer settings between all task pairs, and which aspects of the settings are most predictive of performance change. Comparing the theoretical grounding assessments to predictive heuristics of transfer performance from prior work (e.g., dataset and task embedding similarity)~\cite{poth-etal-2021-pre} and other practical aspects of performance settings (e.g., sample and vocabulary sizes), we find that definition granularity and operationalization correspondence are significantly more predictive of transfer performance than these other features. Further, we analyze transfer performance according to the qualitative themes, finding that \textit{direct} empathy tasks more often improved and less often harmed target tasks. 

Our findings empirically confirm the importance of multidimensional operationalizations of empathy in NLP research~\cite{lahnala-etal-2022-acritical}. The lack of transferability, and moreover, the strength of the conceptual and measured/observed dimensionality in empathy tasks in predicting transfer performance highlight an important issue: abstract notions of empathy are insufficient for the development of NLP models. The lack of evidence that existing empathy prediction models are effective for downstream domains and language settings has important implications for developing robust evaluation frameworks for generative empathy models~\cite{lee-etal-2024-comparative} and emotional intelligence in LLMs~\cite{huang2023emotionally,wang2023emotional,Sabour-emobench}.

This paper contributes the first empirical analysis of the effects of construct operationalization on empathy prediction. Our methodology addresses the difficulty of disentangling aspects of construct operationalization from other transfer aspects, and therefore could be applied to investigate operationalizations of other complex constructs studied in NLP. We will publicly release the models developed for this work upon publication, offering additional tools for the analysis and evaluation of empathy in language. By integrating these models into a singular pipeline, we facilitate further investigation into the unique properties they capture, thereby advancing the development of a robust evaluation framework for empathy in language.

\section{Background and Related Work}\label{sec:related_work}

% \mysubsection{Empathy in NLP.} 
The motivations for NLP models of empathy are broad, ranging from developing empathic conversational agents \cite{morris2018towards,rashkin-etal-2019-towards,naous-etal-2021-empathetic}, integration in counselor training tools \cite{tanana2016comparison}, analysis of interactions with support seekers \cite{xiao2015rate}, and investigations of social interactions in online forums \cite{sharma-etal-2020-computational, zhou-jurgens-2020-condolence, lahnala-etal-2021-exploring} which could support ongoing efforts investigating online support platforms, or ``emotional first aid'' \cite{weisberg2023emotional}. Moreover, empathic and emotional support capabilities are increasingly assessed in evaluation frameworks for LLMs~\cite{huang2023emotionally,chang2024survey,manzoor2024can}. Several datasets including \citeposs{rashkin-etal-2019-towards} EmpatheticDialogs for generation, \citeposs{sharma-etal-2020-computational} Epitome, \citeposs{buechel-etal-2018-modeling} NewsReactions, among others, have been introduced for developing models for empathy detection and generation, with much interest focused on the latter. These systems often incorporate a module for sentiment or emotion (e.g., \citeauthor{majumder-etal-2020-mime, rashkin-etal-2019-towards, lin-etal-2019-moel}), and some aim to incorporate common sense reasoning or other grounding knowledge (e.g., \citeauthor{sabour2021cem,li2022knowledge,lee-etal-2024-comparative}).

Despite broad motivations for NLP models of empathy, numerous works have pointed out the limitations due to limited and/or unclear conceptual operationalizations of empathy. Research on detecting or generating ``empathic language,'' often does not specify empathic processes, behaviors, or social expectations of empathic expression \cite{lahnala-etal-2022-acritical}. Research tends to center on emotional empathy, leaving other facets underexplored, such as how cognitive empathic processes may be observed in interactions~\cite{sharma-etal-2020-computational}. Precise definitions and measurement approaches of empathy in language remain open challenges~\cite{shetty2023scoping}; measurements tend to abstract the construct into single ratings, which fails to evaluate important facets of empathy~\cite{lee-etal-2024-comparative}. 

To investigate the impacts of these issues empirically, our study explores the transferability between empathy tasks.
Our study draws on \citet{poth-etal-2021-pre}, which considered heuristics to estimate task transferability between a variety of intermediate and target NLP tasks. They found that the dataset size and task similarity are often reasonable predictors of transfer performance. 
They average sentence-BERT embeddings over a corpus and measure similarity to predict transfer performance~\cite{reimers2019sentence}.
Other works, such as \citet{luo-etal-2022-cogtaskonomy} examine both neural network (computer) models and fMRI data to derive task representations.
\citet{vu-etal-2020-exploring} explore using task embeddings computed from the Fisher information matrix to predict transfer performance, as the information about useful model parameters serves as a ``rich source of knowledge about the task itself.'' 
These works have demonstrated that the similarity of these embeddings between tasks is an effective heuristic for estimating transferability. We use these heuristics in addition to other practical factors that can affect transfer performance (e.g., the number samples, tokens, and data sources) to assess how relevant the theoretical grounding is to transfer performance by comparison.

\section{Dataset and Task Descriptions}

We study 18 regression and classification tasks across nine English-language datasets (Table~\ref{tab:construct_component_ranks}), with various empathy constructs underlying the dataset collection and task definitions. We outline them here, and provide details in Appendix~\ref{sec:e_transfer_task_data}.

\mysubsection{Social Media Datasets.} The Motivational Interviewing (MI) \cite{welivita-pu-2022-curating}, Condolence \cite{zhou-jurgens-2020-condolence}, and Epitome (EPIT) \cite{sharma-etal-2020-computational} datasets are sourced from Reddit. MI is annotated with MI counselor behaviors at the sentence level using the aforementioned Motivational Interviewing Treatment Integrity (MITI) scheme \cite{moyers2003motivational} in interactions in mental health forums. Epitome also contains interactions in mental health forums, labeled based on the degree to which certain empathy communication mechanisms are exhibited in peer-supporter responses. The Condolence dataset \cite{zhou-jurgens-2020-condolence} contains exchanges from Reddit forums with empathy ratings based on an appraisal theory of empathy with six dimensions \cite{lamm2007neural,wondra2015appraisal}. The Empathy Hope dataset \cite{yoo-etal-2021-empathy} contains geopolitical tweets from India and Pakistan to classify supportive content that expresses empathy, distress, or solidarity.

\mysubsection{Crowd-sourced Datasets.} The News dataset \cite{buechel-etal-2018-modeling} contains empathetic essay reactions to news articles with first-person empathy measurements, and continuous empathy and distress scores are derived from Batson's Empathic Concern – Personal Distress Scale \cite{batson1987distress}. The Conv dataset \cite{omitaomu2022empathic} contains crowd-sourced conversations between two participants about articles in the News dataset, with turn-level annotations of emotional polarity, emotional intensity, and empathy. The EmpDial dataset \cite{rashkin-etal-2019-towards} contains dialogues grounded on specific emotions, drawing on emotion theories such as Ekman's \cite{ekman1971universals,ekman1992argument} and Plutchik's \cite{plutchik1984emotions}.
EmpDial EI \cite{welivita-pu-2020-taxonomy} and EmpDial QI \cite{svikhnushina-etal-2022-taxonomy} are subsets of the EmpDial dataset annotated according to taxonomies of empathic intents in responses and questions.

\section{Empathy Operationalizations}\label{sec:e_transfer_empathy_constructs}

To study how conceptual operationalizations of empathy tasks impact transfer performance,
we assess empathy tasks within NLP based on core premises underlying their scientific methods, namely, 1) the construct \textit{Definition}, 2) the \textit{correspondence (Link)} of the measurements and observations used to operationalize the construct to the construct itself, and 3) the \textit{Conduciveness} of the data to represent the construct.  We developed an annotation task, defining criteria for these aspects along a 5-point Likert scale and rate each task (discussed in \sref{sec:conceptual_ranking_components}; full instructions and scoring criteria are in Figures \ref{fig:emp_transfer_instructions} and \ref{fig:emp_transfer_annotation_scoring_criteria} in the Appendix).

In addition, we carefully considered and discussed the empathy tasks and identified three broad themes (discussed in \sref{sec:construct_groups}) regarding the relationship between the empathy tasks and their theoretical grounding: 1) \textit{direct} empathy, 2) empathy \textit{abstraction}, and 3) empathy \textit{adjacent} tasks. 
Both the annotation of operationalizations and qualitative categorization into themes enable a qualitative analysis of the impact of each on transfer performance.

\subsection{Rating Conceptual Operationalizations}\label{sec:conceptual_ranking_components}

The rating criteria for the definition and correspondence aspects are informed by the empathy definition and evaluation themes identified by \citet{lahnala-etal-2022-acritical}, and prior work in social psychology~\cite{hall-schwartz-2019-empathy}, which similarly surveyed the diversity of conceptual and operational definitions of empathy and the correspondence between the measurements and construct. The third component relates to contextualizing language behaviors within specific situations or contexts ~\cite{boyd_verbal_2024}, and considering expectations around empathy that may influence the data.

\mysubsection{Construct Definition.}
We assess the granularity of each task's conceptual definition of empathy, considering the number of dimensions, behaviors, and other details about the construct provided by the works. They range from providing \textit{no} conceptual definition to embracing empathy's multidimensionality with descriptions of many dimensions, factors, and characteristics of empathic experiences, interactions, language, and conversational behaviors. 

\mysubsection{Operational Correspondence (Link).}
We assess the correspondence of the operationalization of empathy to the defined construct (referred to as \textit{Link} for short) by considering their measurements and observations. For the NLP context, these are typically reflected by the task labels; how directly are the task labels connected to the defined aspects of the underlying construct? Higher ratings reflect tasks that cover several possible empathic behaviors or indicators of the empathy construct. In contrast, lower ratings reflect tasks with labels that are an abstraction of the empathy construct or no labels relating to the construct are involved. As an example, the development of the \textit{Condolence} dataset is based on a fine-grained construct, an \textit{appraisal theory} of empathy \cite{lamm2007neural,wondra2015appraisal} which specifies six perspective dimensions along which an observer can appraise a target's situation. However, the task labels are singular ratings of empathy that abstract the underlying construct based on trained annotators' assessment of how the empathic observer's appraisals align with the support-seeking target by considering the six dimensions at once. Thus, as shown in Tables~\ref{tab:empathy_transfer_annotations} and \ref{tab:construct_component_ranks}, \textit{Condolence} is rated highly by definition but low by link.

\mysubsection{Conduciveness.}
We assess the Conduciveness of the \textit{language scenario} to capture empathic processes in the interactions; i.e., how much could one expect \textit{empathy} to be observable in the language? However, any natural interaction between humans could represent empathic processes \cite{debnath-2023-critical-empatheticdialogues}. 

Thus, we base the ratings on plausible expectations or assumptions that can be made about the scenario, considering aspects like the social norms of the scenario and the data collection methods. For example, we may consider whether a certain level of common ground between participants can be assumed based on experimental controls. Social norm considerations deal with the nature of the language or conversation scenario; for instance, empathy is a social norm for mental health support and therapy conversations, but it is not an expectation of scientific articles. For example, both annotators rate \textit{Conv Empathy} highly based on aspects of the experimental design (see Table~\ref{tab:empathy_transfer_annotations}). Before having a conversation with each other, two participants independently performed the same tasks intended to induce empathic processes/experiences. First, they read the same inherently emotional news article, which grounds their knowledge of the conversation topic; then, they are asked to describe their empathic experience in a short essay. These facets of the pre-conversation task imply the participants begin the conversation with common ground and that they have performed more deliberative empathic processes beyond initial emotional reactions, which we consider conducive to cognitive empathy.

\mysubsection{Annotation Task.} Two authors of this study completed the task and are knowledgeable about empathy constructs within and outside the NLP field. All ratings collected from the annotation task are reported in Table~\ref{tab:empathy_transfer_annotations} in the Appendix.
Table \ref{tab:annotation_agreement_emp_transfer} shows the inter-rater agreement measured by Krippendorff's $\alpha$ and the Spearman rank correlation coefficients $r$ and $p$-values between the annotators on each aspect. 
The agreement and correlation measures are highest on the Definition and Link. 
The higher agreements on Definition and Link versus the lower agreement on Conduciveness are likely affected by relatively more objectivity in the case of Definition and Link. In contrast, Conduciveness is less defined and more subjective to interpretation regarding several aspects that could describe a ``language scenario.'' 
Figure~\ref{fig:empathy_transfer_rating_diffs} shows the degree of difference between annotator ratings on each aspect. We observe that the Conduciveness ratings more often differ by one point; in four cases, they differ by two points. Specific disagreements can be observed in Table~\ref{tab:empathy_transfer_annotations}.

\subsection{Themes of Theoretical Grounding}\label{sec:construct_groups}

Table~\ref{tab:construct_component_ranks} in the Appendix shows the characterization of each task according to the themes described in this section.

\mysubsection{\textit{Direct empathy tasks}} generally have fine-grained construct definitions that link directly to the task labels. This group includes, for instance, MI Behavior, which directly labels counseling behaviors defined by the MITI construct, and EmpDial QInt and EI, which involve the labels of the empathic intents that \textit{comprise} their construct, directly at the dialogue act level. This group also includes the News Empathy and Distress \cite{buechel-etal-2018-modeling}, though News Distress borders on belonging to the next group, \textit{empathy abstraction}. The labels are ratings that capture two facets of an empathy construct, \textit{distress} and \textit{empathy} (empathic concern) are the primary facets they target, so we include them here. Furthermore, the ratings are empirically derived by a \textit{multi-item} scale that measures the empathic observers' internal empathic experience upon reading empathy-inducing articles; thus, we consider them more directly connected to the construct than a third person's abstractive assessment. Similarly, Conv Empathy measures the empathy construct with the same data but through the assessment of a conversation, rather than essay response.

\mysubsection{\textit{Empathy abstraction tasks}} range in granularity of the construct definitions but involve abstraction of the underlying construct, for instance, by binarizing or rating the construct more broadly. This includes the \textit{Condolence} task, which abstracts a fine-grained construct as discussed in Section~\ref{sec:conceptual_ranking_components}, and the \textit{MI Adherent} task, which is a binary abstraction of a set of counseling behaviors related but not explicitly connected to a set of empathic process components. Empathy Hope is included here because it contains labels of supportive content, which is broad and may contain empathy. The three Epitome tasks do separate cognitive and emotional aspects of empathy, but rely on third-person annotations of the text that use strong/weak/none, so we consider it as an abstraction of the expression of each aspect.

\mysubsection{\textit{Empathy-adjacent tasks}} as reflected by the name are those that involve predicting labels that reflect concepts that \textit{relate} to empathy, rather than labels connected to the empathy construct. Naturally, these
have a lower level of empathy construct granularity and Link to empathy construct, but are useful aspects to analyze in empathic interactions, such as emotions and emotion intensity reflected in EmpDial Emo, Conv EmoPol, Conv EmoInt and News Emotion, and dialogue roles and acts reflected in EmpDial Role and EmpDial QAct.

\section{Experiments: Empathy Task Transferal}\label{sec:e_transfer_experiment_setup}

Our study aims to identify the ways that conceptual operationalizations of empathy impact transferability between models. Our experimental paradigm shown in Figure~\ref{fig:paper-fig} is to compare models baseline \textit{target task} models to models tuned first on an \textit{intermediate task} and adapted for the \textit{target} task. Testing each of the 18 empathy tasks as an intermediate for each of the other 17 tasks as targets provides a large space for examining how factors of the transfer settings relate to performance differences from the baselines. 

Full fine-tuning for this number of experiments would require substantial computational resources and energy~\cite{wang-etal-2023-energy}. Therefore, we use the parameter-efficient adapter-tuning strategy. Adapters are a lightweight tuning strategy in which the pre-trained model parameters are frozen, and only the weights of new layers injected into the model are updated when training for a downstream task; this strategy has often matched the performance of full fine-tuning~\cite{houlsby2019parameter}. Furthermore, this is the same approach used by \citet{poth-etal-2021-pre} for exploring heuristics for estimating task transferability between numerous NLP tasks, which informs our work. 

All models have a Transformer-based architecture~\cite{vaswani2017attention} with adapter modules tuned for the empathy tasks. We use RoBERTa-base \cite{liu2019roberta} as the pre-trained Transformer. We tested several types of adapters during a hyperparameter search, finding best outcomes from single bottleneck adapters~\cite{pfeiffer-etal-2020-mad} followed by LoRA~\cite{hu2021lora}, and therefore use the bottleneck adapters across all experiments. Full training and hyperparameter search details are provided in Appendix~\ref{sec:appendix_training_details}. We encode the prior post or turn and the target utterance for each dataset that includes context (all but News and Empathy Hope).

\mysubsection{Baselines.} 
We obtain baseline predictions for each target task by training a task adapter with a prediction head and obtaining predictions from this model on the test split. We performed a hyperparameter search and used the best configuration for each task, shown in Table~\ref{tab:hyperparams}.

\mysubsection{Empathy-to-Empathy task transfer.} The 18 task adapters trained for the target task baselines are subsequently used in the transfer experiments as \textit{intermediate task} adapters. To adapt them for the remaining 17 target tasks, we remove the prediction head from the intermediate adapter, and add a new task adapter and prediction head for the target task, using the stacked composition setup from \citet{pfeiffer-etal-2020-mad}. We train this composition for the target task and use the resulting model for inference on the target test dataset.  

\mysubsection{Results Overview.} The F1 scores for the classification tasks are provided in Table~\ref{tab:classification_f1} and the Pearson $r$ scores for regression tasks are in Table~\ref{tab:regression_pearson}. For each intermediate-to-target pair, we compute the percent change in performance compared to the baselines. The values for the classification target tasks are shown in a heatmap in Figure~\ref{fig:heatmap_transfer_performances_classification} and for the regression target tasks in Figure~\ref{fig:heatmap_transfer_performances_regression}. Statistically significant performance gains across tasks are relatively rare in our experimental results, indicating little is gained from intermediate empathy task tuning. The following sections are in-depth examinations of the results with respect to our RQ.

\begin{figure}[t]
    \centering
    \input{figs/permutation_feature_bar}
    \caption{Feature importances in regression fit to improvement over baseline calculated as the difference in $R^2$ when permuting the feature.}
    \label{fig:feature_importance}
\end{figure}
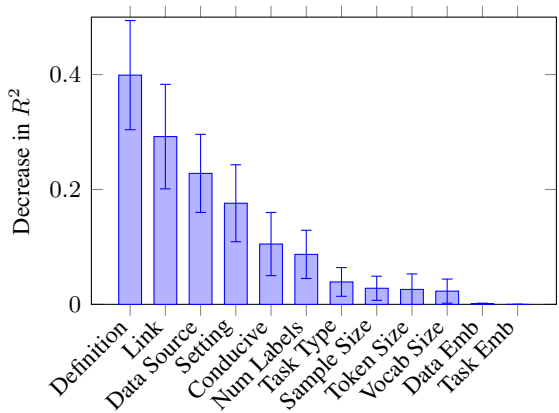

\subsection{Features Impacting Transferability}\label{sec:e_transfer_results}

To understand \textit{how} conceptual operationalizations of empathy impact transfer performance, we consider the importance of the definition granularity, operational correspondence, and data conduciveness (\sref{sec:conceptual_ranking_components}) in predicting the performance change between the baseline and transfer models, in comparison to other features of the transfer settings. 

\mysubsection{Basic Features:} For each intermediate-target task pair, we consider whether the following increases, decreases, or remains the same: the i) number of data samples, ii) number of tokens, iii) number of unique tokens (vocabulary), and (iv) number of labels. Then, we consider (v) the task types--whether the transfer goes from classification to classification or regression, and vice versa; (vi) the source and target tasks' data source (e.g., Reddit-to-Twitter); and (vii) language setting (e.g., conversation, social media, or essay).

\mysubsection{Transferability Heuristics:} We also consider the similarities between the pairs' (viii) dataset embeddings and (ix) task embeddings, which were found be effective at estimating the transferability between NLP tasks \cite{vu-etal-2020-exploring,poth-etal-2021-pre,sileo-moens-2022-analysis}, with higher similarities being associated with better transfer performance.
Following \citet{poth-etal-2021-pre}'s approach, we use the \texttt{all-mpnet-base-v2} model to compute Sentence BERT (SBERT) embeddings over each dataset and subsequently the cosine similarity between each dataset's embedding.
We follow the task embedding similarity aggregation from previous work \cite{vu-etal-2020-exploring}, which combines BERT components using reciprocal rank fusion \cite{cormack2009reciprocal}. 
Heatmaps of the similarities are shown in Figures \ref{fig:heatmap_sbert} and \ref{fig:task_emb_sim_rank}.

\mysubsection{Construct Features:} These features are the differences between the intermediate and target task ratings on the annotated aspects.

\mysubsection{Quantifying Feature Importance:}
We fit a support vector regression model~\cite{awad2015support} on these features to predict the percentage improvement by the transfer models over the baselines. We measure feature importance using the permutation approach, whereby each feature is separately permuted across instances, and the impact on $R^2$ is measured~\cite{breiman2001random}. We permute 50 times for each of the ten models trained on random train/test splits of the 306 trials.

Figure~\ref{fig:feature_importance} shows the feature importances averaged across permutations.\footnote{Values are provided in Table~\ref{tab:feature_importance}.} Definition granularity and operational correspondence (link) are the most predictive of transfer performance, followed by the data source and language setting. Conduciveness and the remaining  basic features are relatively insignificant compared to these features. The heuristic features (data and task emb) were \textit{not} predictive.

\subsection{Qualitative Themes and Performance}\label{sec:qual_themes}

\input{resources/performance_bar_charts2}
The themes outlined in \sref{sec:construct_groups} provide researchers a quick and intuitive way to characterize the theoretical grounding of their empathy tasks: does the task \textit{directly} predict certain components of the empathy construct, predict an \textit{abstract} representation of the construct, or some other construct related or \textit{adjacent} to empathy?
Grouping tasks by these themes, we perform a qualitative analysis of how the theoretical groundings transfer to other ones. 

Figure~\ref{fig:transfer_performance_construct_groups} (top) shows the count of adapters that significantly\footnote{Significance indicates p-value $<0.05$ provided by a non-parametric bootstrap test with resampling.} outperformed and underperformed the task baselines aggregated by these themes. Direct empathy tasks as intermediates are the most frequent to outperform the baseline, but only for direct empathy and empathy adjacent to target tasks. Intermediate tuning does not benefit any abstract tasks. The transfer learning models often significantly \textit{harm} the performance, which occurs least frequently with direct intermediate tasks. The direct empathy tasks are also the most frequent targets to \textit{gain} from intermediate tuning.  

\subsection{Case Study: Limited Data}\label{sec:limited_data}

A motivation for studying empathy task transferability is the potential to utilize empathy models trained on existing resources to increase their scale when training data is limited, and support the construction of new resources. For example, transferability has critical value for supporting technologies for health care like clinician communication training, where annotations are costly and data is sensitive~\cite{tanana2019development,imel2017technology}. 
While the previous section showed little value for \textit{improving} target task performance, could intermediate tuning with \textit{limited training data} at least achieve \textit{similar} results as using no intermediate tuning but the \textit{full training data}? 

To explore this, we tested intermediate tasks in models that use \textit{half} the amount of target task data for target task tuning and compared the performance differences to the target task-only baseline developed on the complete training set. For these results shown in Figure~\ref{fig:transfer_performance_construct_groups} (bottom), we are interested in the rate of significant losses compared to the rate of insignificant differences (or significant improvements), as this indicates whether the transfer effectively enables similar performance to a scenario in which we had twice the training data. 
As intermediate tasks, the task themes have similar rates of insignificant differences or improvements, only direct empathy tasks showing more benefits to abstract target tasks. However, the direct tasks as the target task are much less likely to benefit, shown also by significant harms.

\section{Discussion}

\emph{Conceptual operationalizations have practical impacts.} We examined the theoretical grounding of empathy tasks according to the definition granularity, the correspondence between operationalization and defined components, and the conduciveness of the data to reflect empathy. Analyzing these features in a model fit to the change in performance from baseline to transfer model revealed the significant influence conceptual operationalizations have over transfer performance. This underscores the importance of well-defined, multidimensional constructs and measurements/observed that directly correspond to them (\sref{sec:e_transfer_results}). 

\emph{Constructs are not equally helpful or harmful.}
We explored how tasks function as intermediates and targets based on the themes defined in \sref{sec:qual_themes}, and whether using the transfer models could support limited data settings (\sref{sec:limited_data}). While few transfer models overall significantly improve performance, abstraction target tasks are never improved, indicating that different theoretical empathy constructs are unable to map meaningful properties to abstract ones. Conversely, abstract and adjacent tasks may map to specific components of direct tasks. However, these observations are overshadowed by the much higher rates of performance harms observed for all themes as intermediates and targets, occurring slightly less often from direct intermediates.  
The direct tasks had the most frequent significant improvement and least performance losses compared to baselines fine-tuned only on the target task. 
The limited data setting demonstrated direct empathy had less to gain and more to lose from intermediate transfer. Small differences between the groups may suggest that direct tasks are more likely to effectively support resource expansion and construction, but there is not substantial evidence. 

\emph{NLP needs precise, multidimensional empathy constructs.}
The main takeaway of our findings is that fine-grained definitions operationalized via measurable or observable characteristics in language for a construct such as empathy, which is not directly measurable, are necessary for more reliable representations. The tasks with better transferability may be more beneficial, for instance, when deciding which training data to supply to the training of an empathetic model, but compelling evidence of the benefits is yet to be shown.
Further researchers of NLP empathy modeling should aim for well-defined, multidimensional constructs, and using measurements/observations that directly correspond to the specific components of the construct. Future work could draw inspiration from approaches in psychometrics that utilize multi-item instruments to measure latent constructs \cite{el2020measure}, as was done in developing the \textit{News} and \textit{Conv} tasks \cite{buechel-etal-2018-modeling,omitaomu2022empathic}. 

Empathy constructs studied in the field of psychology commonly involve broad components such as cognitive and emotional empathy, yielding a variety of measurement instruments depending on what facet of empathy is the target of study~\cite{cuff2016empathy}. Some measures may estimate internal empathic experiences of an empathic observer (e.g., Batson's Empathic Concern – Personal Distress Scale \cite{batson1987distress} used by \citealp{buechel-etal-2018-modeling}), while others may characterize empathy in dialog with discourse frameworks (e.g.,  \citealp{pounds2011empathy,velasco2022analysis}). 
While research on empathy in psychology and other fields could provide foundation for NLP research, these fields face many of the same operationalization challenges \cite{coll2017we,hall-schwartz-2019-empathy}. 
\citeposs{hall-schwartz-2019-empathy} survey of psychology literature highlights a prevalent lack of consistency in definitions and measurements, validations of operationalizations, and frequently weak correspondence between construct definition and measurement. While they also observed the rates of stating conceptual definitions and referring to empathy's multidimensionality increasing over time, they argue that the application of the term ``empathy'' could be refined to lower-level constructs and in many cases the term is inappropriate in a scientific sense for their research objective.

Ultimately, as empathy is not directly measurable, operationalizing requires defining multiple indicators and characteristics that can be measured, observed, and tested. As \citeauthor{hall-schwartz-2019-empathy} put it, ``The continued vague use of the term empathy to characterize a wide range of different methods and definitions can only dilute the value of scholarship.'' 
We therefore invite further discussions regarding the question of whether the construct we intend to study is best described as ``empathy'' as opposed to a lower-level construct: What is the specific use for models that can detect or generate empathy? If we scrutinize the empathy construct against the purposes and goals of specific systems, are there more specific and operationalizable constructs that are more appropriate?

\section{Conclusion}

As interest in systems that model empathy grows, it is crucial to address the limitations of NLP research on empathy of whether the conceptual operationalizations lead to reliable models of empathy in language.
This work leveraged an intermediate-task transfer experiment paradigm to investigate transferability of 18 empathy-related tasks in NLP research. We analyzed the performance on the basis of each task's conceptual operationalization of empathy to answer the research question of how the theoretical grounding of empathy tasks impacts empathy modeling. The critical finding is that the \textit{granularity} of empathy definitions \textit{and} how directly the measurements and observations in the data \textit{correspond} to the defined components are significant factors of predicting transfer performance. Future researchers should carefully construct empathy modeling tasks according to well-defined constructs that are measurable or observable in the language data.

\section*{Limitations}

For this study, we chose widely used datasets to provide insight into the properties of resources that researchers are currently exploring, as well as a recently introduced large-scale dataset that differs more in terms of the type of tasks. Further work may explore transferable properties from other empathy and empathy-related datasets and task types, such as generation. 

We note that we introduced three construct groups through deliberation with the authors of this paper. This process involved a few hours of discussion of each datasets qualities and measurements. This process is a qualitative research approach that not easily reproducible, but nonetheless provides a framework for thinking about empathy datasets in future work. 
Quantitatively, we showed that our annotation process can be used to arrive at a similar separation of groups using the average of annotated aspects (see Table~\ref{tab:construct_component_ranks}, however these three numbers by themselves may not capture the full picture. This is a starting point in a larger discussion about how to categorize the constructs of empathy and better understand their impact on downstream task performance. Future work should seek more concrete validation of the construct groups through refining the annotation process.

Our study only contains English data and, thus, only captures properties of how empathy is expressed in English. We expect that there are socio-variations in how empathy is expressed and perceived, not only across languages but also across different English speakers, contexts, and relationships between speakers. Future work could explore transfer learning to support empathy modeling in multi-lingual spaces. There are a variety of empathy resources for other languages, such as Arabic \cite{naous-etal-2021-empathetic}, Italian \cite{alam2018annotating,sanguinetti-etal-2020-annotating}, Japanese \cite{ito-etal-2020-relation}, German \cite{wambsganss-etal-2021-supporting}, and Chinese \cite{sun-etal-2021-psyqa}. This work could explore zero-shot or few-shot cross-lingual transfer to languages without annotated empathy resources, as was done by \citet{pfeiffer-etal-2020-mad} for other NLP tasks.
In our experiments, we test transferring from one task to another in a stacked composition adapter setup. Further research may consider combinations of multiple-task adapters and multi-task learning.

\bibliography{main}
\bibliographystyle{acl_natbib}

\appendix

\section{Dataset and Task Descriptions}\label{sec:e_transfer_task_data}

This appendix provides the details of each dataset and task. Table~\ref{tab:dataset_size} shows the size of datasets in samples, vocabulary size, and tokens. Tables~\ref{tab:labels_4} through \ref{tab:labels_1} contain splits and label distributions.

\begin{table}[h]
    \centering
    \small
\begin{tabular}{lrrr}
\toprule
                  \textbf{Dataset}   &   \textbf{Samples} &   \textbf{$|V|$} &     \textbf{Total Tokens} \\
\midrule
 Condolence &      1,004 & 15,139 & 353,951           \\
 Conv &     11,176 & 11,040 &      3,755,038 \\
 EmpDial     &     76,494 & 26,054 &      2,642,492 \\
 EmpDial EI      &     53,414 & 26,447 &      5,043,441 \\
 EmpQT         &     20,201 & 16,703 &      1,602,514 \\
 Empathy \& Hope          &      1,282 &  6,461 &  31,415           \\
 Epitome                   &      3,023 & 11,882 & 324,084           \\
 MI &     15,645 & 14,876 &      3,344,011 \\
 News Stories &      2,700 &  6,324 & 228,764           \\
\bottomrule
\end{tabular}

    \caption{Number of samples, size of the vocabulary ($|V|$), and total number of tokens in each dataset.}
    \label{tab:dataset_size}
\end{table}

 \subsection{Condolence} 

 The Condolence Empathy dataset contains expressions of condolences exchanged on Reddit forums annotated with empathy ratings from 1.0 to 5.0 \cite{zhou-jurgens-2020-condolence}. Two annotators were trained on an \textit{appraisal theory} of empathy \cite{lamm2007neural,wondra2015appraisal}, which specifies six perspective dimensions that an observer may appraise of the target: (1) pleasantness, (2) anticipated effort, (3) situational control, (4) who was responsible for the situation (self or other), (5) attentional activity, and (6) certainty about the situation or its aftermath. The annotators assessed the observers' condolences in response to targets' expressing their situation, basing their single rating on consideration of these dimensions. The label distributions are shown in Table \ref{tab:labels_4}.

 \subsection{News Stories (News)}\label{sec:news_dataset_description}
 
 The News dataset, introduced by \citet{buechel-etal-2018-modeling}, contains empathetic essay reactions to news articles with first-person empathy measurements collected by the following approach: First, the participants read a news article; second, they completed the Batson Empathic Concern -- Personal Distress Scale \cite{batson1987distress}, a multi-item scale for measuring first-person empathy, and continuous \textit{empathy} and \textit{distress} scores are derived empirically from this scale; and third, they wrote 300-800 character essays about the article. By associating the Batson empathy and distress scores with the text, the dataset offers a pairing of text-to-empathy scores of the actual text writer rather than annotated by a third-person observer. Given these essays, we perform the tasks of predicting the empathy and distress scores with regression models (News Empathy and News Distress). The label distributions are in Table~\ref{tab:labels_4}.

\citet{tafreshi-etal-2021-wassa} extended the News dataset to include person-level demographic and personality information and additional emotion labels. The emotion labels are based on Ekman's six basic emotions (\textit{sadness, joy, disgust, surprise, anger,} and \textit{fear}) \cite{ekman1971universals}, with added \textit{neutral} and \textit{hope} labels. We perform a classification task (News Emotion) to predict these emotion labels for the essay. The News Emotion label distributions are in Table~\ref{tab:labels_3}.
 
 \subsection{Empathic Conversations (Conv)}\label{sec:conv_dataset_description}

 The Conv dataset \cite{omitaomu2022empathic,barriere-etal-2023-findings} contains crowd-sourced conversations between two participants about articles in the News dataset described above. Before the conversation, each participant underwent the same process described for the News dataset. The conversations contain turn-level annotations of emotional polarity--(1) positive, (2) neutral, or (3) negative; emotional intensity--(1) no emotion, (2) weak emotion, (3) moderately strong emotion; (4) very strong emotion; or (5) extremely strong emotion, and empathy, as the degree to which the speaker puts themself in the shoes of the suffering victim--(1) not at all; (2) a little bit; (3) moderately; (4) quite a lot; and (5) extremely. Third-person observers of the conversation provided these annotations. Given the turn and prior dialogue context, we perform three separate regression tasks of predicting each of these assigned scores on each turn (Conv EmoPol, Conv EmoInt, and Conv Empathy). The label distributions for all Conv tasks are in Table \ref{tab:labels_4}.

 \subsection{Empathetic Dialogues (EmpDial)} 
 
 The EmpDial dataset \cite{rashkin-etal-2019-towards} contains dialogues based on emotional experiences and is widely used in NLP for fine-tuning generative empathy models. The speaker (target) first writes about a situation in which they experienced a given emotion. Then, they have a conversation with a listener (observer), in which they tell the story of the situation. The EmpDial conversations are accompanied by the grounding emotion and the situation written by the target in the public dataset. The dataset does not have empathy labels specifically, so we define two other turn-level tasks. First is role prediction (EmpDial Role), distinguishing between the speaker, who initiates the conversation based on a personal emotional situation, and the Listener, who responds without knowing the original emotion label or situation description. The second task is to predict the grounding emotion (EmpDial Emo). The model is provided the turn and prior dialogue context for both tasks. The label distributions for EmpDial Emo and EmpDial Role are in Table~\ref{tab:labels_1} and Table~\ref{tab:labels_3}, respectively.
 
 Other researchers extended the dataset with empathy labels; we use the \textbf{empathic response intents} (EmpDial EI) \cite{welivita-pu-2020-taxonomy} and the \textbf{empathic question intents} (EmpDial QI) \cite{svikhnushina-etal-2022-taxonomy} datasets described below, which are subsets of the original EmpDial dataset.

 \subsection{Empathic Intents (EmpDial EI)} 
 
 EmpDial EI \cite{welivita-pu-2020-taxonomy} is a subset of EmpDial developed by Welivita and Pu (2020). It has nine different categorical labels of empathic intents:  1) questioning, 2) acknowledging, 3) neutral, 4) agreeing, 5) sympathizing, 6) encouraging, 7) suggesting, 8) consoling, and 9) wishing. While Welivita and Pu \cite{welivita-pu-2020-taxonomy} did not develop the scheme on a specific existing empathy construct, their work describes several existing theories from psychology and neuroscience which may inform the scheme, including Zillman's (2008) social emotion definition \cite{zillman-2008-empathy-social}, \textit{simulation theory}  \cite{gordon-robert-1992}, \textit{theory-theory} \cite{gopnik2012reconstructing}, and Singer and Klimecki's (2014) distinction between empathy and compassion \cite{singer2014empathy}. The nine intent labels are applied at the sentence level in turns of the EmpDial dataset. We perform the task of predicting the empathic intent labels at the sentence level, with the prior dialogue context provided as input. The EmpDial EI. label distributions are in Table~\ref{tab:labels_3}.

\subsection{Empathic Question Taxonomy (EmpDial QInt and EmpDial QAct)} 

 After observing that \textit{questioning} was the most frequent empathic intent among the prior subset of EmpDial, \citet{svikhnushina-etal-2022-taxonomy} developed a fine-grained scheme that focuses on the role of questions in empathic conversations, with two different label sets. One set of labels categorizes nine question \textit{acts} (QAct) resembling dialogue acts: 1) Ask about antecedent, 2) Suggest a solution, 3) Request information, 4) Ask about consequence, 5) Positive rhetoric, 6) Negative rhetoric, 7) Suggest a reason, 8) Ask for confirmation, and 9) Irony. The other set focuses on twelve categories of empathic intents underlying questions (QInt) in empathetic conversations: 1) Express concern, 2) Express interest, 3) Moralize speaker, 4) Sympathize, 5) Amplify joy, 6) Amplify excitement, 7) Support, 8) De-escalate, 9) Offer relief, 10) Amplify pride, 11) Motivate, and 12) Pass judgement. According to \citet{svikhnushina-etal-2022-taxonomy}, the coding scheme is informed by prior question classification schemes and reference to the principles of emotional regulation~\cite{gross2013handbook}. Their motivation for fine-grained analyses of questions in empathic interactions draws from other findings from psychology (e.g., \citealp{huang2017doesn}) and linguistics (e.g., \citealp{freed1994form}; \citealp{enfield_questionresponse_2010}) regarding their social role. On this dataset, we perform a task for each label set to predict the labels at the sentence level; the model is provided with the full turn of the sentence and the prior dialogue context. The EmpDial Empathic Intents and Question Intents datasets contain manual and automatic labels using models trained on the manual labels. We train the source adapters on the manually labeled samples in our study. The label distributions for each task are in Table~\ref{tab:labels_2}.

\subsection{Motivational Interviewing (MI)} 

The MI dataset \cite{welivita-pu-2022-curating} contains Reddit interactions between support seekers and support providers curated from several mental health-related subreddits. The support provider turns are expert-annotated using the MITI coding scheme~\cite{moyers2003motivational}, which are counselor behavior categories. Thus, the underlying construct of empathy is drawn from the foundational work on Motivational Interviewing, a counseling style often used for behavior change therapy \cite{miller2012motivational}.
We perform two tasks on this data: a binary classification of MI-adherent and non-adherent behaviors (MI Adherent) and a multi-label classification of fine-grained behaviors (MI Behavior). MI Behavior has fourteen possible categorical labels applied to sentences of support-provider responses. The MI adherent and non-adherent classes are subsets of the fourteen behaviors, which align with the principles of motivational interviewing (adherent) or not (non-adherent). The MI adherent behaviors include 1) \textit{affirm}, 2) \textit{give information}, 3) \textit{complex reflection}, 4) \textit{support}, 5) \textit{closed question}, 6) \textit{emphasize autonomy}, 7) \textit{simple reflection}, 8) \textit{advise with permission} and 9) \textit{open question}. The non-adherent behaviors include 10) \textit{advise without permission}, 11) \textit{self-disclosure}, 12) \textit{direct}, 13) \textit{warn}, and 14) \textit{confront}. 
MI adherent behaviors relate to empathic behaviors and expectations \cite{moyers2013low,lord2015more,perez-rosas-etal-2017-understanding}, especially in the \textit{reflection} behaviors where counselors express their empathic understanding of what the client is saying \cite{mccambridge2011fidelity,pollak2011physician}. The label distributions for MI Behavior and MI Adherent are in Table~\ref{tab:labels_1} and Table~\ref{tab:labels_3}, respectively.

\subsection{Epitome}\label{sec:epitome_description}

The Epitome dataset \cite{sharma-etal-2020-computational} encompasses dialogue exchanges between mental health support seekers and support providers from Reddit. 
Empathy is treated as three discernible forms of \textit{expressed} empathy, including cognitive and emotional components \cite{cuff2016empathy,selman1980growth}: 1) Emotional Reactions (ER), 2) Interpretations (IP), and 3) Explorations (EX), each providing a unique perspective on empathetic responses. 
As defined by the authors, ERs involve emotional expressions such as warmth, compassion, and concern; IPs communicate an inferred understanding of feelings and experiences; EXs represent efforts to improve understanding by investigating unstated feelings and experiences in the post. We perform the tasks of predicting the empathy level for each category on the supporter responses: no communication (0), weak communication (1), and strong communication (2). 
The model is provided with the support seeker context prior to the supporter's turn; note, however, a study by \citet{lee-etal-2023-empathy} found that models tend to ignore the speaker context, which we do not analyze in this work. 
The label distributions for each task are in Table~\ref{tab:labels_3}.

\subsection{Empathy and Hope} 

Empathy Hope is a dataset of geopolitical Tweets from India and Pakistan \cite{yoo-etal-2021-empathy}. \citeposs{yoo-etal-2021-empathy} work focused on the health crisis emerging from the COVID-19 pandemic and the expressions of solidarity across country lines. They developed the dataset for detecting \textit{supportive} content, which the dataset creators define as content expressing empathy, distress, or solidarity. This dataset's empathy construct is based on the News dataset's construct \cite{buechel-etal-2018-modeling}, defining empathy as a ``warm, tender, and compassionate feeling for a suffering entity,'' and distress as ``a self-focused, negative affective state that occurs when one feels upset due to witnessing an entity's suffering or need.'' They apply a model developed on \citeposs{buechel-etal-2018-modeling} data of empathy and distress, in addition to a hope speech classifier \cite{palakodety2020hope}, which they use as signals for detecting annotated labels of supportive versus not-supportive content. We perform the task of predicting the manual labels of supportive versus not-supportive content. The label distributions are shown in Table~\ref{tab:labels_3}.

\begin{figure*}[h!]
    \centering
    
\begin{tcolorbox}[colback=blue!5,colframe=blue!75!black,title=Instructions for Empathy Operationalization Annotation Task]
% \small
\textbf{\textsc{Context}}

We are investigating how empathy constructs are operationalized for NLP research via measurable or observable characteristics in language. We aim to understand whether the various operationalizations of the construct are convergent and valid, capturing aspects of the targeted construct. In order to characterize the operationalizations, we are collecting human annotated ratings for 1) the level of granularity of the construction definition, 2) the correspondence of the measurements or observations in language to the construct definition, and 3) the Conduciveness of the language scenario to represent empathic processes in the language data. The annotation task requires understanding these sources' scientific methods and experimental design, namely, how each paper described the construct of empathy, their empathy measurement or observation methods, and the experimental setting from which the data was collected or obtained. Thus, background knowledge of empathy constructs in psychology and the landscape of empathy research in NLP is necessary; prior knowledge of each dataset and task is beneficial.        

\vspace{12pt}

\textbf{\textsc{Instructions}}

We will provide descriptions of 18 empathy language tasks from 9 different datasets and references to the original articles where they were proposed. Note that some tasks are defined here and not in the original paper, which we added to support our study. You will be asked to rate these three aspects on a 5-point Likert scale:

Each task/dataset will be annotated according to three aspects on a 5-point Likert Scale:

\begin{enumerate}[nosep]
    \item Definition: How granular (coarse/high level to fine/more detailed) is the empathy construct or theory that grounds the data development or task definition?
    \item Link/Correspondence: To what degree do the measurement and/or observation approaches correspond to the construct's defined components?
    \item Conduciveness: How does the language scenario/communication context$^*$ influence your expectation of observing the empathy construct in the data, considering aspects such as data collection methods (survey, interview, observation, experiment, etc.), annotation criteria, and social norms of the context?  
\end{enumerate}

$^*$\textit{Language scenario} refers to properties of the corpus of communications, such as: What form of communication does the data reflect? Is the corpus a collection of conversations--if so, are they asynchronous, goal-oriented, etc? Is it a collection of interactions in online forums--if so, what purpose do the online forums serve, or are there various purposes? Is it a collection of messages in isolation from an interaction (e.g., an essay)? What was the data collection method (e.g., experiment, survey, observation)? Does the corpus center on a particular domain?
\tcblower
% \small
Next, we will provide the scoring criteria for rating these aspects on the 5-point Likert scale (see Figure~\ref{fig:emp_transfer_annotation_scoring_criteria}).

\end{tcolorbox}
    \caption{Instructions for the empathy operationalization annotation task.}
    \label{fig:emp_transfer_instructions}
\end{figure*}

\begin{figure*}[h!]
    \centering
    
\begin{tcolorbox}[colback=blue!5,colframe=blue!75!black,title=Scoring Criteria for Empathy Operationalization Categorization]

% \small

\textbf{Definition.} 
% \vspace{5pt}
\begin{enumerate}[nosep]
    \item The empathy concept is neither defined nor described nor is there a referenced source from which a definition is drawn.
    \item The empathy concept defined or referenced bypasses its multidimensionality, e.g., by focusing on a single aspect without relating it to or discussing other possible aspects, by being abstract, simplified, or vague.
    \item The empathy concept defined or referenced acknowledges more than one dimension of empathy but leaves these dimensions abstract, e.g., by merely referring to emotional and cognitive empathy.
    \item The empathy concept has a fairly fine granularity. It comprises more than one dimension of empathy, such as cognitive and emotional empathy, with high-level descriptions of how they emerge in language or are experienced and perceived.
    \item The empathy concept has a very fine granularity in the dimensions/factors/aspects hypothesized to comprise empathic experiences or interactions by including several concrete dimensions of empathic experiences, behaviors, or interactions.

\end{enumerate}
% \tcblower
% \vspace{8pt}
\textbf{Link/Correspondence.} 
% \vspace{5pt}
\begin{enumerate}[nosep]
    \item The measurements or observations have a \textit{very weak} correspondence to the defined construct. They may capture aspects of empathy or related phenomena, but not of the defined empathy construct by any direct or indirect approach.
    \item The measurements or observations have a \textit{weak} correspondence to the defined construct. They seem intended to reflect the empathy construct, but the connection is unclear, or a high level of abstraction likely interferes with capturing what is intended.
    \item The measurements or observations \textit{somewhat} correspond to the defined construct. They clearly reflect the defined construct, but there is a fair amount of abstraction; some elements of the construct are missing, or the methods may not be reliable.
    \item The measurements or observations correspond \textit{fairly well} to the defined construct. There may be some abstraction, but they are systematically derived based on several aspects of the construct with moderately reliable methods. 
    \item The measurements or observations correspond \textit{very well} to the defined construct; they directly correspond to the individual facets of the construct, and the methods seem highly appropriate and precise for measuring what is intended.
\end{enumerate}
% \tcblower
% \vspace{8pt}
\textbf{Conduciveness.} 
% \vspace{5pt}
\begin{enumerate}[nosep]
    \item Aspects of the scenario \textit{significantly} lower my expectations of observing empathy.
    \item Aspects of the scenario \textit{somewhat} lower my expectations of observing empathy.
    \item The properties of the scenario do not affect my expectations of observing empathy.
    \item Aspects of the scenario \textit{somewhat} increase my expectations of observing empathy.
    \item Aspects of the scenario \textit{significantly} increase my expectations of observing empathy.
\end{enumerate}

\end{tcolorbox}
    \caption{Criteria provided to the annotators for scoring each aspect of the empathy construct along a 5-point Likert scale.}
    \label{fig:emp_transfer_annotation_scoring_criteria}
\end{figure*}

\begin{table*}[]
    \centering
    \begin{tabular}{|l|cc|cc|cc|}
\hline
\textbf{Task}         &   \multicolumn{2}{c|}{\textbf{Definition}} &  \multicolumn{2}{c}{\textbf{Link}} &   \multicolumn{2}{|c|}{\textbf{Conducive}} \\
       &   A &   B &  A &   B &   A &   B \\
\hline
 Condolence   &   5 &   5 &   2 &   2 &   3 &   4 \\
 Conv EmoInt  &   3 &   3 &   2 &   1 &   4 &   3 \\
 Conv EmoPol  &   3 &   3 &   2 &   2 &   4 &   5 \\
 Conv Empathy &   5 &   5 &   4 &   3 &   5 &   5 \\
 EmpDial EI.  &   4 &   5 &   5 &   5 &   2 &   4 \\
 EmpDial Emo  &   2 &   2 &   1 &   1 &   1 &   3 \\
 EmpDial Role &   1 &   2 &   1 &   1 &   1 &   2 \\
 Empathy Hope &   2 &   3 &   3 &   2 &   1 &   1 \\
 Epitome ER   &   3 &   3 &   3 &   3 &   3 &   4 \\
 Epitome EX   &   3 &   3 &   3 &   3 &   3 &   4 \\
 Epitome IP   &   3 &   3 &   3 &   3 &   3 &   4 \\
 EmpDial QAct &   3 &   3 &   5 &   3 &   2 &   4 \\
 EmpDial QInt &   4 &   4 &   5 &   5 &   2 &   4 \\
 MI Adherent  &   2 &   3 &   3 &   2 &   3 &   3 \\
 MI Behavior  &   3 &   4 &   4 &   4 &   3 &   4 \\
 News Distress     &   4 &   5 &   4 &   4 &   4 &   4 \\
 News Emotion &   3 &   3 &   3 &   2 &   4 &   3 \\
 News Empathy &   5 &   5 &   5 &   4 &   4 &   4 \\
\hline
\end{tabular}
    \caption{\textbf{Results from the construct aspect rating annotation task}. Here shows the 5-point Likert scale construct aspect ratings from each annotator A and B for each  (Definition, Link, Conduciveness) on each task. Generally, the annotator ratings differ by 0 or 1 point; refer to Figure~\ref{fig:empathy_transfer_rating_diffs} for difference counts by annotator pair.}
    \label{tab:empathy_transfer_annotations}
\end{table*}

\begin{table*}[t]
    \centering
    \small
    \begin{tabular}{ccccccccc}
\toprule
  \textbf{Metric}   &   \multicolumn{2}{c}{\textbf{Definition}} &   \multicolumn{2}{c}{\textbf{Link}} &   \multicolumn{2}{c}{\textbf{Conduciveness}} & \multicolumn{2}{c}{\textbf{Overall}}\\
\midrule
Krippendorff's $\alpha$ &   \multicolumn{2}{c}{0.86} &   \multicolumn{2}{c}{0.83} &  \multicolumn{2}{c}{0.46} & \multicolumn{2}{c}{0.72} \\
\midrule

 & $r$ & $p$-value & $r$ & $p$-value & $r$ & $p$-value & $r$ & $p$-value \\
Spearman's $r$ &    0.91 &      0.00 &   0.89 &       0.00 &            0.49 &            0.04 &  0.70 &      0.00 \\   
\bottomrule
\end{tabular}
    \caption{\textbf{Krippendorff's $\alpha$ values for measuring agreement between the annotators, and Spearman's rank correlation coefficients $r$ between the Likert scores provided by the annotators.} The agreement is high on Definition and Link, but there is more disagreement on Conduciveness. The correlations are statistically significant ($p$-value $\leq$ 0.05), with positive coefficients across each aspect.}
    \label{tab:annotation_agreement_emp_transfer}
\end{table*}

\begin{table*}[]
    \centering
\begin{tabular}{|l|c|c|c|}
\hline
 \textbf{Task}         &   \textbf{Epoch} &   \textbf{Batch Size} &   \textbf{Learning Rate} \\
\hline
 Condolence   &       8 &           32 &         0.00028 \\
 Conv EmoInt  &       7 &           54 &         0.00083 \\
 Conv EmoPol  &       8 &           32 &         0.00093 \\
 Conv Empathy &       8 &           31 &         0.00083 \\
 EmpDial EI.  &      11 &           44 &         0.00073 \\
 EmpDial Emo  &       9 &           18 &         0.00069 \\
 EmpDial Role &      11 &           12 &         0.00086 \\
 Empathy Hope &       7 &           47 &         0.00064 \\
 Epitome ER   &       7 &           42 &         0.00041 \\
 Epitome EX   &      10 &           34 &         0.00090 \\
 Epitome IP   &       9 &           22 &         0.00078 \\
 EmpDial QAct. &       8 &           24 &         0.00056 \\
 EmpDial QInt. &      10 &           52 &         0.00078 \\
 MI Adherent  &       8 &           10 &         0.00076 \\
 MI Behavior  &      10 &           30 &         0.00048 \\
 News Distress     &       6 &           30 &         0.00087 \\
 News Emotion &       6 &           15 &         0.00043 \\
 News Empathy &      11 &           14 &         0.00045 \\
\hline
\end{tabular}
    \caption{Hyperparameter configurations for the best adapters for each empathy task on the validation sets.}
    \label{tab:hyperparams}
\end{table*}

\begin{table}[t!]
\small
    \centering
\begin{tabular}{lccccc}
\toprule
              &   Def. &   Link &   Cond. &  \makecell{Mean} \\%& \makecell{Construct\\Group} \\
\midrule
\textbf{Direct} & & & & \\%& \\
 News Empathy &         5.0 &   4.5 &        4.0 &           4.50 \\ %& \multirow{6}{*}{\rotatebox[origin=c]{0}{Direct}} \\
 Conv Empathy &         5.0 &   3.5 &        5.0 &           4.50 \\%& \\
 EmpDial EI.  &         4.5 &   5.0 &        3.0 &           4.17 \\%& \\
 News Distress     &         4.5 &   4.0 &        4.0 &           4.17 \\%& \\
 EmpDial QInt.&         4.0 &   5.0 &        3.0 &           4.00 \\%& \\
 MI Behavior  &         3.5 &   4.0 &        3.5 &           3.67 \\%& \\
 \midrule
 \textbf{Abstract} &&&& \\
 Condolence   &         5.0 &   2.0 &        3.5 &           3.50 \\%& \multirow{6}{*}{\rotatebox[origin=c]{0}{Abstract}} \\
 Epitome EX   &         3.0 &   3.0 &        3.5 &           3.17 \\%&\\
 Epitome ER   &         3.0 &   3.0 &        3.5 &           3.17 \\%&\\
 Epitome IP   &         3.0 &   3.0 &        3.5 &           3.17 \\%&\\
 Empathy Hope* &         2.5 &   2.5 &        1.0 &           2.00 \\%&\\
 MI Adherent*  &         2.5 &   2.5 &        3.0 &           2.67 \\%&\\
 \midrule
 \textbf{Adjacent} &&&& \\
 EmpDial QAct.* &         3.0 &   4.0 &        3.0 &           3.33 \\%& \multirow{6}{*}{\rotatebox[origin=c]{0}{Adjacent}}\\
 Conv EmoPol*  &         3.0 &   2.0 &        4.5 &           3.17 \\%&\\
 News Emotion &         3.0 &   2.5 &        3.5 &           3.00 \\%&   \\
 Conv EmoInt  &         3.0 &   1.5 &        3.5 &           2.67 \\%&\\
 EmpDial Emo  &         2.0 &   1.0 &        2.0 &           1.67 \\%&\\
 EmpDial Role &         1.5 &   1.0 &        1.5 &           1.33 \\%&\\
\bottomrule
\end{tabular}
    \caption{\textbf{Themes of Theoretical Grounding.} The means of the annotators' ratings per task are shown on the right. We see that the mean ratings almost separate the datasets into the three distinct groups, which map to the authors deliberative categorization. Deviations are marked with *. 
    }
    \label{tab:construct_component_ranks}
\end{table}

\section{Annotations and Construct Groups}

In this section, we present Table~\ref{tab:construct_component_ranks} showing the separate themes and their corresponding annotations for definition, link, and conduciveness. We find that the cut-offs for the means are close to linearly separating the three groups, with a clear cutoff between direct and abstract, but a less clear distinction between abstract and adjacent.

\section{Training Details}\label{sec:appendix_training_details}

We ran a hyperparameter search using all empathy tasks; the best hyperparameter configurations for each empathy task are shown in Table~\ref{tab:hyperparams}. Models were run for ten epochs, with learning rates in the range of $1e^{-3}$ to $1e^{-6}$, batch size in the range of 8 to 64, taking the validation set performance from the best epoch. We considered the adapter model type as a hyperparameter. We tested the bottleneck adapter~\cite{pfeiffer-etal-2020-mad}, which adds layers after the feed-forward block in each transformer layer; the double bottleneck~\cite{houlsby2019parameter}, which adds layers both before and after the feed-forward block; prefix tuning~\cite{li-liang-2021-prefix}, which adds trainable prefix parameters to the keys and values in the attention heads; prompt tuning~\cite{lester-etal-2021-power}, which instead appends tunable tokens to the input text; compacter~\cite{karimi2021compacter}, which adds parameterized hypercomplex multiplication layers~\cite{zhang2021beyond} in place of the feed-forward adapter layers; and LoRA~\cite{hu2021lora}, which instead uses low-rank decomposition matrices in the attention layers. 
After 500 trials, we found that bottleneck adapters and LoRA outperformed other methods in all cases. We used the best batch size and learning rate for each task and trained these three adapters on all empathy tasks. The single bottleneck adapter performed best most of the time across task validation sets. Since we intend to stack adapters and compare task embeddings, we used this adapter type for all subsequent experiments.

\input{resources/classification_task_heatmap}
\input{resources/regression_task_heatmap}

\begin{figure}[t]
    \centering
    \includegraphics[width=\linewidth]{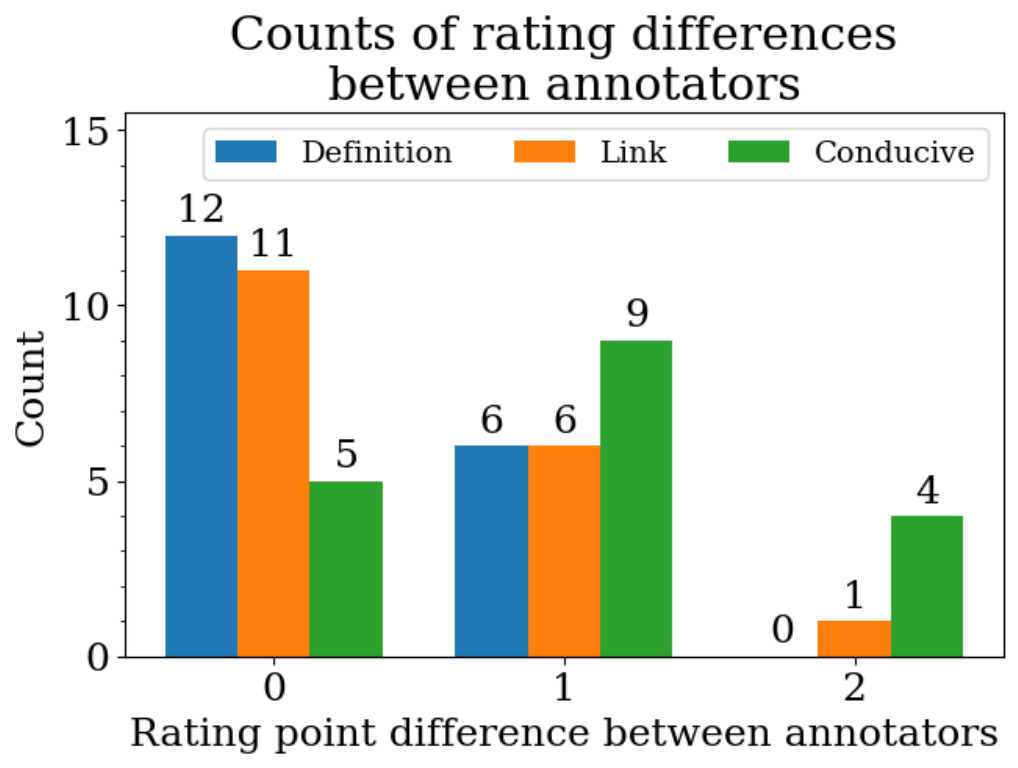}
    \caption{\textbf{Counts of differences in ratings by the annotators on each construct aspect.} For Definition and Link, the annotators most frequently differ by zero or one point. However, Conduciveness has a larger degree of disagreement, and the annotators more frequently differ by one point.}
    \label{fig:empathy_transfer_rating_diffs}
\end{figure}

\begin{table}[t]
    \centering
    \small
\begin{tabular}{lc}
\toprule
 \textbf{Feature}          & \textbf{Importance}        \\
\midrule

 C1 Definition    & $0.399 \pm 0.095$  \\
 C2 Link          & $0.292 \pm 0.091$  \\
 Data Source      & $0.228 \pm 0.068$  \\
 Language Setting & $0.176 \pm 0.067$  \\
 C3 Conducive     & $0.105 \pm 0.055$  \\
 Number Labels    & $0.087 \pm 0.042$  \\
 Task Type        & $0.039 \pm 0.025$  \\
 Sample Size      & $0.028 \pm 0.021$  \\
 Token Size       & $0.026 \pm 0.027$  \\
 Vocab Size       & $0.023 \pm 0.021$  \\
 Data Emb Sim     & $0.001 \pm 0.001$  \\
 Task Emb Sim     & $0.000 \pm 0.000$ \\
\bottomrule
\end{tabular}
    \caption{Feature importances in regression fit to improvement over baseline calculated as the difference in $R^2$ when permuting the feature.}
    \label{tab:feature_importance}
\end{table}

\input{resources/similar_grid}

\begin{figure*}[t]
    \centering
    \scriptsize
    \input{resources/task_emb_sim}
    \caption{Heatmap of the task embedding similarity rankings. These are computed by ranking the cosine similarities between specific model layers of each task and then computing RRF over those rankings.}
    \label{fig:task_emb_sim_rank}
\end{figure*}

\begin{table*}[]
    \centering
    \small
\begin{tabular}{lcccccccccccc}
\toprule
 & \multicolumn{3}{c}{EmpDial} & Empathy & \multicolumn{3}{c}{Epitome} & \multicolumn{2}{c}{EmpDial} & \multicolumn{2}{c}{MI} & News  \\
              &     \makecell{EI}  &   \makecell{Emo} &   \makecell{Role} &   \makecell{Hope} &   \makecell{ER} &   \makecell{EX} &   \makecell{IP} &   \makecell{QAct} &   \makecell{QInt} &   \makecell{Adherent} &   \makecell{Behavior} &   \makecell{Emotion} \\
\midrule
 Condolence   &          .88 &          .38 &           .94 &           .84 &         .81 &         .76 &         .59 &           .58 &           .38 &          .40 &          .53 &           .67 \\
 Conv EmoInt  &          .88 &          .38 &           .95 &           .89 &         .79 &         .74 &         .56 &           .56 &           .41 &          .81 &          .51 &           .74 \\
 Conv EmoPol  &          .88 &          .38 &           .94 &           .86 &         .81 &         .76 &         .56 &           .60 &           .39 &          .82 &          .51 &           .66 \\
 Conv Empathy &          .88 &          .38 &           .95 &           .87 &         .80 &         .76 &         .59 &           .57 &           .36 &          .81 &          .52 &           .72 \\
 EmpDial EI.  &          \textbf{\underline{.89}} &          .38 &           .94 &           .87 &         .80 &         .75 &         .60 &           .59 &           .37 &          .83 &          .52 &           .80 \\
 EmpDial Emo  &          .88 &          \textbf{\underline{.39}} &           .94 &           .84 &         .78 &         .70 &         .61 &           .57 &           .38 &          \textbf{.83} &          .51 &           \textbf{.87} \\
 EmpDial Role &          .85 &          .36 &           \underline{.95} &           .87 &         .78 &         .77 &         .55 &           .53 &           .38 &          .82 &          .49 &           .53 \\
 Empathy Hope &          .88 &          .38 &           .95 &           \textbf{\underline{.92}} &         .78 &         .74 &         .55 &           .59 &           .40 &          .82 &          .50 &           .59 \\
 Epitome ER   &          .88 &          .38 &           .95 &           .87 &         \textbf{\underline{.82}} &         .77 &         .56 &           .59 &           .38 &          .83 &          .51 &           .74 \\
 Epitome EX   &          .87 &          .04 &           .94 &           .86 &         .82 &         \underline{.79} &         .56 &           .59 &           .39 &          .83 &          .50 &           .40 \\
 Epitome IP   &          .88 &          .38 &           \textbf{.95} &           .83 &         .82 &         .78 &         \textbf{\underline{.63}} &           .58 &           .36 &          .82 &          .51 &           .68 \\
 QAct. Manual &          .88 &          .38 &           .94 &           .86 &         .81 &         .76 &         .59 &           \textbf{\underline{.61}} &           .38 &          .83 &          .50 &           .76 \\
 QInt. Manual &          .88 &          .38 &           .94 &           .86 &         .80 &         .75 &         .60 &           .57 &           \underline{.39} &          .83 &          .52 &           .79 \\
 MI Adherent  &          .88 &          .37 &           .94 &           .89 &         .81 &         .72 &         .60 &           .58 &           .36 &          \underline{.83} &          .50 &           .54 \\
 MI Behavior  &          .88 &          .38 &           .95 &           .89 &         .80 &         .76 &         .56 &           .58 &           .38 &          .82 &          \textbf{\underline{.53}} &           .79 \\
 News Dis     &          .88 &          .38 &           .94 &           \textbf{.92 }&         .80 &         \textbf{.84} &         .56 &           .57 &           .36 &          .83 &          .50 &           .50 \\
 News Emotion &          .88 &          .38 &           .95 &           .83 &         .77 &         .74 &         .61 &           .58 &           \textbf{.43} &          .82 &          .52 &           \underline{.73} \\
 News Empathy &          .87 &          .38 &           .94 &           .91 &         .80 &         .75 &         .56 &           .58 &           .38 &          .81 &          .52 &           .62 \\
\bottomrule
\end{tabular}
    \caption{F1 scores for each target task (columns). The baseline scores (i.e., source is the same as target) are underlined, and the best result is in bold.}
    \label{tab:classification_f1}
\end{table*}

\begin{table*}[]
    \centering
    \small
\begin{tabular}{lrrrrrr}
\toprule
              &   Condolence &   Conv EmoInt &   Conv EmoPol &   Conv Empathy &   News Dis &   News Empathy \\
\midrule
 Condolence   &         \underline{.43} &          .71 &          .73 &           .65 &       .72 &           .91 \\
 Conv EmoInt  &         .42 &          \underline{.74} &          .73 &           .64 &       .75 &           .93 \\
 Conv EmoPol  &         .38 &          .74 &          \underline{.73} &           .67 &       .68 &           .90 \\
 Conv Empathy &         \textbf{.49} &          .73 &          .70 &           \underline{.64} &       .75 &           .91 \\
 EmpDial EI.  &         .42 &          .73 &          .72 &           .66 &       .77 &           .93 \\
 EmpDial Emo  &         .26 &          .72 &          .72 &           .64 &       .79 &           \textbf{.94} \\
 EmpDial Role &         .32 &          .72 &          .73 &           .62 &       .71 &           .90 \\
 Empathy Hope &         .37 &          .73 &          \textbf{.75} &           .63 &       .78 &           .92 \\
 Epitome ER   &         .40 &          .73 &          .73 &           .63 &       .77 &           .92 \\
 Epitome EX   &         .36 &          .72 &          .73 &           \textbf{.68} &       .69 &           .91 \\
 Epitome IP   &         .40 &          .72 &          .73 &           .67 &       .77 &           .92 \\
 QAct. Manual &         .25 &          .73 &          .73 &           .64 &       .75 &           .92 \\
 QInt. Manual &         .34 &          .73 &          .73 &           .67 &       .76 &           .94 \\
 MI Adherent  &         .38 &          .74 &          .69 &           .66 &       .69 &           .90 \\
 MI Behavior  &         .40 &          \textbf{.76} &          .73 &           .66 &       .77 &           .91 \\
 News Dis     &         .37 &          .72 &          .74 &           .64 &       \underline{.71} &           .90 \\
 News Emotion &         .38 &          .71 &          .71 &           .65 &       .78 &           .92 \\
 News Empathy &         .31 &          .71 &          .73 &           .63 &       \textbf{.85} &           \underline{.92} \\
\bottomrule
\end{tabular}
    \caption{Pearsonr scores for each target task (columns). The baseline scores (i.e., source is the same as target) are underlined, and the best result is in bold.}
    \label{tab:regression_pearson}
\end{table*}

\input{resources/label_distribution_tables}

\end{document}

%% file: figs/permutation_feature_bar.tex
\begin{tikzpicture}
    \begin{axis}[
    width=\linewidth,
    height=.7*\linewidth,
        ybar,
        symbolic x coords={Definition, Link, Data Source, Setting, Conducive, Num Labels, Task Type, Sample Size, Token Size, Vocab Size, Data Emb, Task Emb},
        xtick=data,
        ylabel={Decrease in $R^2$},
        ymin=0, ymax=0.5,
        bar width=0.3cm,
        x tick label style={rotate=45, anchor=east},
        error bars/.cd,
        ylabel style={font=\footnotesize},
        y tick label style={rotate=0, font=\footnotesize},
        x tick label style={rotate=0, font=\footnotesize},
        enlarge x limits=0.1,
    ]
    \addplot+ [
        ybar,
        error bars/.cd,
        y dir=both, 
        y explicit,
    ] plot coordinates {
        (Definition, 0.399) +- (0,0.095)
        (Link, 0.292) +- (0,0.091)
        (Data Source, 0.228) +- (0,0.068)
        (Setting, 0.176) +- (0,0.067)
        (Conducive, 0.105) +- (0,0.055)
        (Num Labels, 0.087) +- (0,0.042)
        (Task Type, 0.039) +- (0,0.025)
        (Sample Size, 0.028) +- (0,0.021)
        (Token Size, 0.026) +- (0,0.027)
        (Vocab Size, 0.023) +- (0,0.021)
        (Data Emb, 0.001) +- (0,0.001)
        (Task Emb, 0) +- (0,0)
    };
    \end{axis}
\end{tikzpicture}

%% file: resources/performance_bar_charts2.tex
\begin{figure}[t]
    \centering
    \small
    \begin{tikzpicture}

        % First plot
        \begin{scope}[xshift=-2.5cm,yshift=3.5cm]
            \begin{axis}[
                width  = 4.0cm,
                height = 4cm, %4.5cm, %5cm,
                enlarge y limits=0.25,
                xbar=2*\pgflinewidth,
                bar width= 6pt,%7pt, %9pt,
                nodes near coords,
                xlabel={Intermed. Task},
                ylabel={Target Task},
                ylabel style={font=\small},
                xmajorgrids = true,
                symbolic y coords={Direct, Abstract, Adjacent},
                y tick label style={rotate=45, font=\small},
                ytick = data,
                scaled y ticks = false,
                xmin=0,
                xmax=25,
                ]
                \addplot[fill=white,postaction={pattern=north west lines},area legend] coordinates {(8,Direct) (0,Abstract) (3,Adjacent)};
                \addplot[fill=lightgray,postaction={pattern=crosshatch},area legend] coordinates {(6,Direct) (0,Abstract) (0,Adjacent)};
                \addplot[style={fill=darkgray},area legend] coordinates {(6,Direct) (0,Abstract) (2,Adjacent)};
            \end{axis}
        \end{scope}

        % Second plot
        \begin{scope}[yshift=3.5cm]
            \begin{axis}[
                width  = 4.0cm,
                height = 4cm, %4.5cm, %5cm,
                enlarge y limits=0.25,
                xbar=2*\pgflinewidth,
                bar width= 6pt,%7pt, %9pt,
                nodes near coords,
                xlabel={Intermed. Task},
                ylabel style={font=\small},
                xmajorgrids = true,
                symbolic y coords={Direct, Abstract, Adjacent},
                y tick label style={rotate=0, font=\small},
                ytick = data,
                scaled y ticks = false,
                xmin=0,
                xmax=25,
                yticklabel=\empty,
                ]
                \addplot[fill=white,postaction={pattern=north west lines},area legend] coordinates {(11,Direct) (11,Abstract) (13,Adjacent)};
                \addplot[fill=lightgray,postaction={pattern=crosshatch},area legend] coordinates {(14,Direct) (11,Abstract) (16,Adjacent)};
                \addplot[style={fill=darkgray},area legend] coordinates {(14,Direct) (17,Abstract) (14,Adjacent)};
            \end{axis}
        \end{scope}

        % Third plot
        \begin{scope}[yshift=0cm]
            \begin{axis}[
                width  = 4.0cm,
                height = 4cm, %4.5cm, %5cm,
                enlarge y limits=0.25,
                xbar=2*\pgflinewidth,
                bar width= 6pt,%7pt, %9pt,
                nodes near coords,
                xlabel={Intermed. Task},
                ylabel style={font=\small},
                xmajorgrids = true,
                symbolic y coords={Direct, Abstract, Adjacent},
                y tick label style={rotate=0, font=\small},
                ytick = data,
                scaled y ticks = false,
                xmin=0,
                xmax=35,
                yticklabel=\empty,
                ]
                \addplot[fill=white,postaction={pattern=north west lines},area legend] coordinates {(20,Direct) (13,Abstract) (15,Adjacent)};
                \addplot[fill=lightgray,postaction={pattern=crosshatch},area legend] coordinates {(26,Direct) (14,Abstract) (19,Adjacent)};
                \addplot[style={fill=darkgray},area legend] coordinates {(25,Direct) (18,Abstract) (12,Adjacent)};
            \end{axis}
        \end{scope}

        % Fourth plot
        \begin{scope}[xshift=-2.5cm]
            \begin{axis}[
                width  = 4.0cm,
                height = 4cm, %4.5cm, %5cm,
                enlarge y limits=0.25,
                xbar=2*\pgflinewidth,
                bar width= 6pt,%7pt, %9pt,
                nodes near coords,
                xlabel={Intermed. Task},
                ylabel={Target Task},
                ylabel style={font=\small},
                xmajorgrids = true,
                symbolic y coords={Direct, Abstract, Adjacent},
                y tick label style={rotate=45, font=\small},
                ytick = data,
                scaled y ticks = false,
                xmin=0,
                xmax=35
                ]
                \addplot[fill=white,postaction={pattern=north west lines},area legend] coordinates {(9,Direct) (22,Abstract) (16,Adjacent)};
                \addplot[fill=lightgray,postaction={pattern=crosshatch},area legend] coordinates {(9,Direct) (16,Abstract) (17,Adjacent)};
                \addplot[style={fill=darkgray},area legend] coordinates {(8,Direct) (18,Abstract) (16,Adjacent)};
            \end{axis}
        \end{scope}

        % Legend at the bottom
        \begin{scope}[xshift=-4.5cm, yshift=-6.5cm] % Adjust the xshift and yshift to position the legend
            \begin{axis}[
                hide axis,
                xmin=0,
                xmax=1,
                ymin=0,
                ymax=1,
                legend columns=3,
                legend style={
                    draw=black, % Adds a border around the legend
                    column sep=0cm,
                    /tikz/every even column/.append style={column sep=0.5cm}
                }
                ]
                \addplot[fill=white,postaction={pattern=north west lines},area legend] coordinates {(0,0)};
                \addplot[fill=lightgray,postaction={pattern=crosshatch},area legend] coordinates {(0,0)};
                \addplot[style={fill=darkgray},area legend] coordinates {(0,0)};
                \legend{Direct, Abstract, Adjacent}
            \end{axis}
        \end{scope}

    \end{tikzpicture}
    \vspace{-5cm} % Adjust the negative value to reduce the space between the TikZ picture and the caption
    \caption{\textbf{Transfer performance by theme.} Top: Significant improvement (left) and significant harm (right) counts for full data. Bottom: Insignificant difference or significant improvement (left) and significant harm (right) counts for limited data. 
    % Note the difference in the x-axis limits between the top and bottom figures.
    }
    \label{fig:transfer_performance_construct_groups}
\end{figure}

%% file: resources/classification_task_heatmap.tex
\begin{figure*}[t!]
\scriptsize
\centering
\begin{tikzpicture}
    \begin{axis}[enlargelimits=false,width=12cm,height=10cm,colorbar,colormap/Greys,
        xtick={0,1,2,3,4,5,6,7,8,9,10,11,12,13}, % {0,1,2,3,4,5,6}
        xticklabels={EmpDial EI.,EmpDial Emo,EmpDial Role,Empathy Hope,Epitome ER,Epitome EX,Epitome IP,EmpDial QAct.,EmpDial QInt.,MI Adherent,MI Behavior,News Emotion}, %  {CONV,Buchel,MI,ED,EPI}
        ytick={0,1,2,3,4,5,6,7,8,9,10,11,12,13,14,15,16,17,18,19},
        yticklabels={Condolence,Conv EmoInt,Conv EmoPol,Conv Empathy,EmpDial EI.,EmpDial Emo,EmpDial Role,Empathy Hope,Epitome ER,Epitome EX,Epitome IP,EmpDial QAct.,EmpDial QInt.,MI Adherent,MI Behavior,News Distress,News Emotion,News Empathy}, %{CONV,Buchel,MI,ED,EPI},
        xticklabel style={rotate=90},
        xlabel={Target task},
        ylabel={Source task adapter},
        every node near coord/.append style={yshift=-0.25cm},
        ytick style={draw=none}, xtick style={draw=none}, 
    ]
    \addplot [matrix plot,point meta=explicit, ]
        coordinates {
            % (x,y) position and the [cell value]
            % (x0,y0) [v0] (x1,y0) [v1] ... (xn,y0)

            % (x0,y1) [v0] (x1,y1) [v1] ... (xn,y1)
            % IMPORTANT: make sure there is an empty line between each row
            (0,0) [-0.38] (1,0) [-1.82] (2,0) [-0.51] (3,0) [-8.74] (4,0) [-1.76] (5,0) [-4.35] (6,0) [-5.94] (7,0) [-4.28] (8,0) [-2.71] (9,0) [-51.72] (10,0) [-0.88] (11,0) [-7.77]

                (0,1) [-0.53] (1,1) [-0.81] (2,1) [0.09] (3,1) [-3.43] (4,1) [-4.00] (5,1) [-6.92] (6,1) [-10.69] (7,1) [-8.20] (8,1) [5.01] (9,1) [-3.02] (10,1) [-4.44] (11,1) [1.27]

                (0,2) [-0.95] (1,2) [-2.10] (2,2) [-0.10] (3,2) [-7.01] (4,2) [-1.58] (5,2) [-4.05] (6,2) [-11.79] (7,2) [-1.59] (8,2) [0.07] (9,2) [-1.30] (10,2) [-3.18] (11,2) [-8.58]

                (0,3) [-0.81] (1,3) [-0.71] (2,3) [0.05] (3,3) [-6.13] (4,3) [-2.45] (5,3) [-4.51] (6,3) [-7.40] (7,3) [-6.36] (8,3) [-5.73] (9,3) [-3.43] (10,3) [-2.36] (11,3) [-0.88]

                (0,4) [-35.55] (1,4) [-1.40] (2,4) [-0.21] (3,4) [-5.24] (4,4) [-2.66] (5,4) [-5.16] (6,4) [-5.79] (7,4) [-2.87] (8,4) [-3.02] (9,4) [-0.62] (10,4) [-2.15] (11,4) [9.66]

                (0,5) [-0.61] (1,5) [-35.55] (2,5) [-0.20] (3,5) [-8.62] (4,5) [-5.59] (5,5) [-11.15] (6,5) [-2.72] (7,5) [-5.70] (8,5) [-2.01] (9,5) [0.06] (10,5) [-3.73] (11,5) [19.00]

                (0,6) [-3.93] (1,6) [-7.42] (2,6) [-35.55] (3,6) [-6.00] (4,6) [-5.63] (5,6) [-3.36] (6,6) [-13.24] (7,6) [-12.34] (8,6) [-2.53] (9,6) [-2.05] (10,6) [-7.54] (11,6) [-27.36]

                (0,7) [-1.14] (1,7) [-1.24] (2,7) [0.07] (3,7) [-35.55] (4,7) [-4.91] (5,7) [-6.49] (6,7) [-12.57] (7,7) [-2.63] (8,7) [2.57] (9,7) [-1.43] (10,7) [-6.27] (11,7) [-19.36]

                (0,8) [-0.84] (1,8) [-0.45] (2,8) [0.04] (3,8) [-6.13] (4,8) [-35.55] (5,8) [-2.12] (6,8) [-10.69] (7,8) [-3.30] (8,8) [-2.48] (9,8) [-0.39] (10,8) [-3.97] (11,8) [1.31]

                (0,9) [-1.41] (1,9) [-90.13] (2,9) [-0.31] (3,9) [-6.77] (4,9) [-0.68] (5,9) [-35.55] (6,9) [-11.07] (7,9) [-3.54] (8,9) [1.33] (9,9) [-0.75] (10,9) [-5.21] (11,9) [-44.74]

                (0,10) [-1.06] (1,10) [-2.39] (2,10) [0.22] (3,10) [-9.56] (4,10) [-0.74] (5,10) [-1.45] (6,10) [-35.55] (7,10) [-3.66] (8,10) [-5.56] (9,10) [-1.91] (10,10) [-4.08] (11,10) [-5.97]

                (0,11) [-1.14] (1,11) [-0.56] (2,11) [-0.12] (3,11) [-6.87] (4,11) [-2.24] (5,11) [-3.55] (6,11) [-6.78] (7,11) [-35.55] (8,11) [-1.27] (9,11) [-0.99] (10,11) [-6.65] (11,11) [4.78]

                (0,12) [-1.09] (1,12) [-1.71] (2,12) [-0.11] (3,12) [-7.01] (4,12) [-3.09] (5,12) [-5.56] (6,12) [-4.62] (7,12) [-6.53] (8,12) [-35.55] (9,12) [-0.65] (10,12) [-1.83] (11,12) [8.75]

                (0,13) [-1.13] (1,13) [-4.84] (2,13) [-0.20] (3,13) [-3.38] (4,13) [-1.76] (5,13) [-8.50] (6,13) [-4.66] (7,13) [-5.23] (8,13) [-7.12] (9,13) [-35.55] (10,13) [-6.44] (11,13) [-25.27]

                (0,14) [-0.92] (1,14) [-0.31] (2,14) [0.13] (3,14) [-3.49] (4,14) [-2.31] (5,14) [-4.06] (6,14) [-10.78] (7,14) [-4.38] (8,14) [-2.63] (9,14) [-1.15] (10,14) [-35.55] (11,14) [9.34]

                (0,15) [-1.09] (1,15) [-1.71] (2,15) [-0.21] (3,15) [0.00] (4,15) [-2.39] (5,15) [5.51] (6,15) [-11.85] (7,15) [-6.33] (8,15) [-6.57] (9,15) [-1.04] (10,15) [-6.13] (11,15) [-30.88]

                (0,16) [-0.56] (1,16) [-1.21] (2,16) [-0.09] (3,16) [-9.72] (4,16) [-6.99] (5,16) [-6.49] (6,16) [-3.82] (7,16) [-4.66] (8,16) [10.14] (9,16) [-1.59] (10,16) [-2.26] (11,16) [-35.55]

                (0,17) [-1.64] (1,17) [-1.60] (2,17) [-0.28] (3,17) [-0.85] (4,17) [-2.51] (5,17) [-5.79] (6,17) [-10.76] (7,17) [-5.12] (8,17) [-0.84] (9,17) [-3.19] (10,17) [-2.73] (11,17) [-15.15]
        };
    % white text if value >= .7ish, black if smaller
    % \node[text=white if value >= .7ish else black]
    % for each y, for each x:
    %   \node[text=color] at (axis cs: x,y) {v};
    \node[text=white] at (axis cs: 0,0) {-0.4};
        \node[text=white] at (axis cs: 1,0) {-1.8};
        \node[text=white] at (axis cs: 2,0) {-0.5};
        \node[text=white] at (axis cs: 3,0) {-8.7};
        \node[text=white] at (axis cs: 4,0) {-1.8};
        \node[text=white] at (axis cs: 5,0) {-4.3};
        \node[text=white] at (axis cs: 6,0) {-5.9};
        \node[text=white] at (axis cs: 7,0) {-4.3};
        \node[text=white] at (axis cs: 8,0) {-2.7};
        \node[text=black] at (axis cs: 9,0) {-51.7};
        \node[text=white] at (axis cs: 10,0) {-0.9};
        \node[text=white] at (axis cs: 11,0) {-7.8};
        \node[text=white] at (axis cs: 0,1) {-0.5};
        \node[text=white] at (axis cs: 1,1) {-0.8};
        \node[text=white] at (axis cs: 2,1) {0.1};
        \node[text=white] at (axis cs: 3,1) {-3.4};
        \node[text=white] at (axis cs: 4,1) {-4.0};
        \node[text=white] at (axis cs: 5,1) {-6.9};
        \node[text=white] at (axis cs: 6,1) {-10.7};
        \node[text=white] at (axis cs: 7,1) {-8.2};
        \node[text=white] at (axis cs: 8,1) {5.0};
        \node[text=white] at (axis cs: 9,1) {-3.0};
        \node[text=white] at (axis cs: 10,1) {-4.4};
        \node[text=white] at (axis cs: 11,1) {1.3};
        \node[text=white] at (axis cs: 0,2) {-0.9};
        \node[text=white] at (axis cs: 1,2) {-2.1};
        \node[text=white] at (axis cs: 2,2) {-0.1};
        \node[text=white] at (axis cs: 3,2) {-7.0};
        \node[text=white] at (axis cs: 4,2) {-1.6};
        \node[text=white] at (axis cs: 5,2) {-4.0};
        \node[text=white] at (axis cs: 6,2) {-11.8};
        \node[text=white] at (axis cs: 7,2) {-1.6};
        \node[text=white] at (axis cs: 8,2) {0.1};
        \node[text=white] at (axis cs: 9,2) {-1.3};
        \node[text=white] at (axis cs: 10,2) {-3.2};
        \node[text=white] at (axis cs: 11,2) {-8.6};
        \node[text=white] at (axis cs: 0,3) {-0.8};
        \node[text=white] at (axis cs: 1,3) {-0.7};
        \node[text=white] at (axis cs: 2,3) {0.1};
        \node[text=white] at (axis cs: 3,3) {-6.1};
        \node[text=white] at (axis cs: 4,3) {-2.5};
        \node[text=white] at (axis cs: 5,3) {-4.5};
        \node[text=white] at (axis cs: 6,3) {-7.4};
        \node[text=white] at (axis cs: 7,3) {-6.4};
        \node[text=white] at (axis cs: 8,3) {-5.7};
        \node[text=white] at (axis cs: 9,3) {-3.4};
        \node[text=white] at (axis cs: 10,3) {-2.4};
        \node[text=white] at (axis cs: 11,3) {-0.9};
        \node[text=white] at (axis cs: 0,4) {0.0};
        \node[text=white] at (axis cs: 1,4) {-1.4};
        \node[text=white] at (axis cs: 2,4) {-0.2};
        \node[text=white] at (axis cs: 3,4) {-5.2};
        \node[text=white] at (axis cs: 4,4) {-2.7};
        \node[text=white] at (axis cs: 5,4) {-5.2};
        \node[text=white] at (axis cs: 6,4) {-5.8};
        \node[text=white] at (axis cs: 7,4) {-2.9};
        \node[text=white] at (axis cs: 8,4) {-3.0};
        \node[text=white] at (axis cs: 9,4) {-0.6};
        \node[text=white] at (axis cs: 10,4) {-2.1};
        \node[text=white] at (axis cs: 11,4) {9.7};
        \node[text=white] at (axis cs: 0,5) {-0.6};
        \node[text=white] at (axis cs: 1,5) {0.0};
        \node[text=white] at (axis cs: 2,5) {-0.2};
        \node[text=white] at (axis cs: 3,5) {-8.6};
        \node[text=white] at (axis cs: 4,5) {-5.6};
        \node[text=white] at (axis cs: 5,5) {-11.1};
        \node[text=white] at (axis cs: 6,5) {-2.7};
        \node[text=white] at (axis cs: 7,5) {-5.7};
        \node[text=white] at (axis cs: 8,5) {-2.0};
        \node[text=white] at (axis cs: 9,5) {0.1};
        \node[text=white] at (axis cs: 10,5) {-3.7};
        \node[text=white] at (axis cs: 11,5) {19.0};
        \node[text=white] at (axis cs: 0,6) {-3.9};
        \node[text=white] at (axis cs: 1,6) {-7.4};
        \node[text=white] at (axis cs: 2,6) {0.0};
        \node[text=white] at (axis cs: 3,6) {-6.0};
        \node[text=white] at (axis cs: 4,6) {-5.6};
        \node[text=white] at (axis cs: 5,6) {-3.4};
        \node[text=white] at (axis cs: 6,6) {-13.2};
        \node[text=white] at (axis cs: 7,6) {-12.3};
        \node[text=white] at (axis cs: 8,6) {-2.5};
        \node[text=white] at (axis cs: 9,6) {-2.1};
        \node[text=white] at (axis cs: 10,6) {-7.5};
        \node[text=white] at (axis cs: 11,6) {-27.4};
        \node[text=white] at (axis cs: 0,7) {-1.1};
        \node[text=white] at (axis cs: 1,7) {-1.2};
        \node[text=white] at (axis cs: 2,7) {0.1};
        \node[text=white] at (axis cs: 3,7) {0.0};
        \node[text=white] at (axis cs: 4,7) {-4.9};
        \node[text=white] at (axis cs: 5,7) {-6.5};
        \node[text=white] at (axis cs: 6,7) {-12.6};
        \node[text=white] at (axis cs: 7,7) {-2.6};
        \node[text=white] at (axis cs: 8,7) {2.6};
        \node[text=white] at (axis cs: 9,7) {-1.4};
        \node[text=white] at (axis cs: 10,7) {-6.3};
        \node[text=white] at (axis cs: 11,7) {-19.4};
        \node[text=white] at (axis cs: 0,8) {-0.8};
        \node[text=white] at (axis cs: 1,8) {-0.4};
        \node[text=white] at (axis cs: 2,8) {0.0};
        \node[text=white] at (axis cs: 3,8) {-6.1};
        \node[text=white] at (axis cs: 4,8) {0.0};
        \node[text=white] at (axis cs: 5,8) {-2.1};
        \node[text=white] at (axis cs: 6,8) {-10.7};
        \node[text=white] at (axis cs: 7,8) {-3.3};
        \node[text=white] at (axis cs: 8,8) {-2.5};
        \node[text=white] at (axis cs: 9,8) {-0.4};
        \node[text=white] at (axis cs: 10,8) {-4.0};
        \node[text=white] at (axis cs: 11,8) {1.3};
        \node[text=white] at (axis cs: 0,9) {-1.4};
        \node[text=black] at (axis cs: 1,9) {-90.1};
        \node[text=white] at (axis cs: 2,9) {-0.3};
        \node[text=white] at (axis cs: 3,9) {-6.8};
        \node[text=white] at (axis cs: 4,9) {-0.7};
        \node[text=white] at (axis cs: 5,9) {0.0};
        \node[text=white] at (axis cs: 6,9) {-11.1};
        \node[text=white] at (axis cs: 7,9) {-3.5};
        \node[text=white] at (axis cs: 8,9) {1.3};
        \node[text=white] at (axis cs: 9,9) {-0.7};
        \node[text=white] at (axis cs: 10,9) {-5.2};
        \node[text=black] at (axis cs: 11,9) {-44.7};
        \node[text=white] at (axis cs: 0,10) {-1.1};
        \node[text=white] at (axis cs: 1,10) {-2.4};
        \node[text=white] at (axis cs: 2,10) {0.2};
        \node[text=white] at (axis cs: 3,10) {-9.6};
        \node[text=white] at (axis cs: 4,10) {-0.7};
        \node[text=white] at (axis cs: 5,10) {-1.4};
        \node[text=white] at (axis cs: 6,10) {0.0};
        \node[text=white] at (axis cs: 7,10) {-3.7};
        \node[text=white] at (axis cs: 8,10) {-5.6};
        \node[text=white] at (axis cs: 9,10) {-1.9};
        \node[text=white] at (axis cs: 10,10) {-4.1};
        \node[text=white] at (axis cs: 11,10) {-6.0};
        \node[text=white] at (axis cs: 0,11) {-1.1};
        \node[text=white] at (axis cs: 1,11) {-0.6};
        \node[text=white] at (axis cs: 2,11) {-0.1};
        \node[text=white] at (axis cs: 3,11) {-6.9};
        \node[text=white] at (axis cs: 4,11) {-2.2};
        \node[text=white] at (axis cs: 5,11) {-3.5};
        \node[text=white] at (axis cs: 6,11) {-6.8};
        \node[text=white] at (axis cs: 7,11) {0.0};
        \node[text=white] at (axis cs: 8,11) {-1.3};
        \node[text=white] at (axis cs: 9,11) {-1.0};
        \node[text=white] at (axis cs: 10,11) {-6.7};
        \node[text=white] at (axis cs: 11,11) {4.8};
        \node[text=white] at (axis cs: 0,12) {-1.1};
        \node[text=white] at (axis cs: 1,12) {-1.7};
        \node[text=white] at (axis cs: 2,12) {-0.1};
        \node[text=white] at (axis cs: 3,12) {-7.0};
        \node[text=white] at (axis cs: 4,12) {-3.1};
        \node[text=white] at (axis cs: 5,12) {-5.6};
        \node[text=white] at (axis cs: 6,12) {-4.6};
        \node[text=white] at (axis cs: 7,12) {-6.5};
        \node[text=white] at (axis cs: 8,12) {0.0};
        \node[text=white] at (axis cs: 9,12) {-0.7};
        \node[text=white] at (axis cs: 10,12) {-1.8};
        \node[text=white] at (axis cs: 11,12) {8.8};
        \node[text=white] at (axis cs: 0,13) {-1.1};
        \node[text=white] at (axis cs: 1,13) {-4.8};
        \node[text=white] at (axis cs: 2,13) {-0.2};
        \node[text=white] at (axis cs: 3,13) {-3.4};
        \node[text=white] at (axis cs: 4,13) {-1.8};
        \node[text=white] at (axis cs: 5,13) {-8.5};
        \node[text=white] at (axis cs: 6,13) {-4.7};
        \node[text=white] at (axis cs: 7,13) {-5.2};
        \node[text=white] at (axis cs: 8,13) {-7.1};
        \node[text=white] at (axis cs: 9,13) {0.0};
        \node[text=white] at (axis cs: 10,13) {-6.4};
        \node[text=white] at (axis cs: 11,13) {-25.3};
        \node[text=white] at (axis cs: 0,14) {-0.9};
        \node[text=white] at (axis cs: 1,14) {-0.3};
        \node[text=white] at (axis cs: 2,14) {0.1};
        \node[text=white] at (axis cs: 3,14) {-3.5};
        \node[text=white] at (axis cs: 4,14) {-2.3};
        \node[text=white] at (axis cs: 5,14) {-4.1};
        \node[text=white] at (axis cs: 6,14) {-10.8};
        \node[text=white] at (axis cs: 7,14) {-4.4};
        \node[text=white] at (axis cs: 8,14) {-2.6};
        \node[text=white] at (axis cs: 9,14) {-1.1};
        \node[text=white] at (axis cs: 10,14) {0.0};
        \node[text=white] at (axis cs: 11,14) {9.3};
        \node[text=white] at (axis cs: 0,15) {-1.1};
        \node[text=white] at (axis cs: 1,15) {-1.7};
        \node[text=white] at (axis cs: 2,15) {-0.2};
        \node[text=white] at (axis cs: 3,15) {0.0};
        \node[text=white] at (axis cs: 4,15) {-2.4};
        \node[text=white] at (axis cs: 5,15) {5.5};
        \node[text=white] at (axis cs: 6,15) {-11.8};
        \node[text=white] at (axis cs: 7,15) {-6.3};
        \node[text=white] at (axis cs: 8,15) {-6.6};
        \node[text=white] at (axis cs: 9,15) {-1.0};
        \node[text=white] at (axis cs: 10,15) {-6.1};
        \node[text=white] at (axis cs: 11,15) {-30.9};
        \node[text=white] at (axis cs: 0,16) {-0.6};
        \node[text=white] at (axis cs: 1,16) {-1.2};
        \node[text=white] at (axis cs: 2,16) {-0.1};
        \node[text=white] at (axis cs: 3,16) {-9.7};
        \node[text=white] at (axis cs: 4,16) {-7.0};
        \node[text=white] at (axis cs: 5,16) {-6.5};
        \node[text=white] at (axis cs: 6,16) {-3.8};
        \node[text=white] at (axis cs: 7,16) {-4.7};
        \node[text=white] at (axis cs: 8,16) {10.1};
        \node[text=white] at (axis cs: 9,16) {-1.6};
        \node[text=white] at (axis cs: 10,16) {-2.3};
        \node[text=white] at (axis cs: 11,16) {0.0};
        \node[text=white] at (axis cs: 0,17) {-1.6};
        \node[text=white] at (axis cs: 1,17) {-1.6};
        \node[text=white] at (axis cs: 2,17) {-0.3};
        \node[text=white] at (axis cs: 3,17) {-0.9};
        \node[text=white] at (axis cs: 4,17) {-2.5};
        \node[text=white] at (axis cs: 5,17) {-5.8};
        \node[text=white] at (axis cs: 6,17) {-10.8};
        \node[text=white] at (axis cs: 7,17) {-5.1};
        \node[text=white] at (axis cs: 8,17) {-0.8};
        \node[text=white] at (axis cs: 9,17) {-3.2};
        \node[text=white] at (axis cs: 10,17) {-2.7};
        \node[text=white] at (axis cs: 11,17) {-15.2};

        \fill[white] (-0.5,3.5) rectangle (0.5,4.5);
        \fill[pattern=north west lines, pattern color=black] (-0.5,3.5) rectangle (0.5,4.5);

        \fill[white] (0.5,4.5) rectangle (1.5,5.5);
        \fill[pattern=north west lines, pattern color=black] (0.5,4.5) rectangle (1.5,5.5);

        \fill[white] (1.5,5.5) rectangle (2.5,6.5);
        \fill[pattern=north west lines, pattern color=black] (1.5,5.5) rectangle (2.5,6.5);

        \fill[white] (2.5,6.5) rectangle (3.5,7.5);
        \fill[pattern=north west lines, pattern color=black] (2.5,6.5) rectangle (3.5,7.5);

        \fill[white] (3.5,7.5) rectangle (4.5,8.5);
        \fill[pattern=north west lines, pattern color=black] (3.5,7.5) rectangle (4.5,8.5);

        \fill[white] (4.5,8.5) rectangle (5.5,9.5);
        \fill[pattern=north west lines, pattern color=black] (4.5,8.5) rectangle (5.5,9.5);

        \fill[white] (5.5,9.5) rectangle (6.5,10.5);
        \fill[pattern=north west lines, pattern color=black] (5.5,9.5) rectangle (6.5,10.5);

        \fill[white] (6.5,10.5) rectangle (7.5,11.5);
        \fill[pattern=north west lines, pattern color=black] (6.5,10.5) rectangle (7.5,11.5);

        \fill[white] (7.5,11.5) rectangle (8.5,12.5);
        \fill[pattern=north west lines, pattern color=black] (7.5,11.5) rectangle (8.5,12.5);

        \fill[white] (8.5,12.5) rectangle (9.5,13.5);
        \fill[pattern=north west lines, pattern color=black] (8.5,12.5) rectangle (9.5,13.5);

        \fill[white] (9.5,13.5) rectangle (10.5,14.5);
        \fill[pattern=north west lines, pattern color=black] (9.5,13.5) rectangle (10.5,14.5);

        \fill[white] (10.5,15.5) rectangle (11.5,16.5);
        \fill[pattern=north west lines, pattern color=black] (10.5,15.5) rectangle (11.5,16.5);

    \end{axis}
\end{tikzpicture}

\caption{Classification task heat map of transfer performance of empathy source adapters on empathy target tasks. Values are calculated as the percent improvement of the transfer model over the baseline performance measured by the F1 Score.}
\label{fig:heatmap_transfer_performances_classification}
\end{figure*}

%% file: resources/regression_task_heatmap.tex
\begin{figure}[ht]
\scriptsize
\centering
\begin{tikzpicture}
    \begin{axis}[
        enlargelimits=false,
        width=5.5cm,
        height=10cm,
        colorbar,
        colorbar style={width=2.5mm},
        colormap/Greys,
        xtick={0,1,2,3,4,5,6,7}, 
        xticklabels={Condolence,Conv EmoInt,Conv EmoPol,Conv Empathy,News Distress,News Empathy}, 
        ytick={0,1,2,3,4,5,6,7,8,9,10,11,12,13,14,15,16,17,18,19},
        yticklabels={Condolence,Conv EmoInt,Conv EmoPol,Conv Empathy,EmpDial EI.,EmpDial Emo,EmpDial Role,Empathy Hope,Epitome ER,Epitome EX,Epitome IP,EmpDial QAct.,EmpDial QInt.,MI Adherent,MI Behavior,News Distress,News Emotion,News Empathy}, %{CONV,Buchel,MI,ED,EPI},
        xticklabel style={rotate=90},
        xlabel={Target task},
        ylabel={Source task adapter},
        every node near coord/.append style={yshift=-0.25cm},
        ytick style={draw=none},
        xtick style={draw=none}, 
    ]
    \addplot [matrix plot,point meta=explicit, ]
        coordinates {
            % (x,y) position and the [cell value]
            % (x0,y0) [v0] (x1,y0) [v1] ... (xn,y0)

            % (x0,y1) [v0] (x1,y1) [v1] ... (xn,y1)
            % IMPORTANT: make sure there is an empty line between each row
            (0,0) [0.00] (1,0) [-4.42] (2,0) [0.13] (3,0) [2.40] (4,0) [1.20] (5,0) [-1.04]

                (0,1) [-1.96] (1,1) [0.00] (2,1) [0.51] (3,1) [0.11] (4,1) [6.56] (5,1) [0.78]

                (0,2) [-11.68] (1,2) [0.85] (2,2) [0.00] (3,2) [4.29] (4,2) [-3.54] (5,2) [-2.29]

                (0,3) [14.25] (1,3) [-1.51] (2,3) [-3.90] (3,3) [0.00] (4,3) [5.37] (5,3) [-0.73]

                (0,4) [-3.32] (1,4) [-0.97] (2,4) [-1.49] (3,4) [3.73] (4,4) [9.02] (5,4) [1.62]

                (0,5) [-40.54] (1,5) [-2.92] (2,5) [-1.81] (3,5) [0.39] (4,5) [10.96] (5,5) [2.31]

                (0,6) [-26.23] (1,6) [-2.70] (2,6) [-0.25] (3,6) [-2.38] (4,6) [0.44] (5,6) [-2.00]

                (0,7) [-14.30] (1,7) [-1.18] (2,7) [2.93] (3,7) [-1.95] (4,7) [9.55] (5,7) [0.54]

                (0,8) [-7.69] (1,8) [-1.68] (2,8) [-0.07] (3,8) [-0.95] (4,8) [8.96] (5,8) [0.11]

                (0,9) [-17.50] (1,9) [-2.21] (2,9) [0.02] (3,9) [6.13] (4,9) [-1.89] (5,9) [-1.28]

                (0,10) [-7.64] (1,10) [-2.22] (2,10) [0.60] (3,10) [5.63] (4,10) [8.37] (5,10) [0.28]

                (0,11) [-43.20] (1,11) [-1.35] (2,11) [-0.58] (3,11) [0.34] (4,11) [6.33] (5,11) [0.03]

                (0,12) [-21.00] (1,12) [-1.43] (2,12) [0.01] (3,12) [4.61] (4,12) [7.04] (5,12) [1.75]

                (0,13) [-11.61] (1,13) [0.76] (2,13) [-5.30] (3,13) [3.83] (4,13) [-2.03] (5,13) [-2.29]

                (0,14) [-6.62] (1,14) [2.77] (2,14) [0.75] (3,14) [2.84] (4,14) [8.12] (5,14) [-0.74]

                (0,15) [-14.44] (1,15) [-2.37] (2,15) [1.89] (3,15) [1.00] (4,15) [0.00] (5,15) [-1.65]

                (0,16) [-12.99] (1,16) [-3.92] (2,16) [-2.46] (3,16) [1.10] (4,16) [10.48] (5,16) [0.47]

                (0,17) [-27.81] (1,17) [-3.90] (2,17) [0.50] (3,17) [-1.89] (4,17) [19.81] (5,17) [0.00]
        };
    % white text if value >= .7ish, black if smaller
    % \node[text=white if value >= .7ish else black]
    % for each y, for each x:
    %   \node[text=color] at (axis cs: x,y) {v};
    \node[text=white] at (axis cs: 0,0) {0.0};
        \node[text=white] at (axis cs: 1,0) {-4.4};
        \node[text=white] at (axis cs: 2,0) {0.1};
        \node[text=white] at (axis cs: 3,0) {2.4};
        \node[text=white] at (axis cs: 4,0) {1.2};
        \node[text=white] at (axis cs: 5,0) {-1.0};
        \node[text=white] at (axis cs: 0,1) {-2.0};
        \node[text=white] at (axis cs: 1,1) {0.0};
        \node[text=white] at (axis cs: 2,1) {0.5};
        \node[text=white] at (axis cs: 3,1) {0.1};
        \node[text=white] at (axis cs: 4,1) {6.6};
        \node[text=white] at (axis cs: 5,1) {0.8};
        \node[text=white] at (axis cs: 0,2) {-11.7};
        \node[text=white] at (axis cs: 1,2) {0.9};
        \node[text=white] at (axis cs: 2,2) {0.0};
        \node[text=white] at (axis cs: 3,2) {4.3};
        \node[text=white] at (axis cs: 4,2) {-3.5};
        \node[text=white] at (axis cs: 5,2) {-2.3};
        \node[text=white] at (axis cs: 0,3) {14.2};
        \node[text=white] at (axis cs: 1,3) {-1.5};
        \node[text=white] at (axis cs: 2,3) {-3.9};
        \node[text=white] at (axis cs: 3,3) {0.0};
        \node[text=white] at (axis cs: 4,3) {5.4};
        \node[text=white] at (axis cs: 5,3) {-0.7};
        \node[text=white] at (axis cs: 0,4) {-3.3};
        \node[text=white] at (axis cs: 1,4) {-1.0};
        \node[text=white] at (axis cs: 2,4) {-1.5};
        \node[text=white] at (axis cs: 3,4) {3.7};
        \node[text=white] at (axis cs: 4,4) {9.0};
        \node[text=white] at (axis cs: 5,4) {1.6};
        \node[text=black] at (axis cs: 0,5) {-40.5};
        \node[text=white] at (axis cs: 1,5) {-2.9};
        \node[text=white] at (axis cs: 2,5) {-1.8};
        \node[text=white] at (axis cs: 3,5) {0.4};
        \node[text=white] at (axis cs: 4,5) {11.0};
        \node[text=white] at (axis cs: 5,5) {2.3};
        \node[text=black] at (axis cs: 0,6) {-26.2};
        \node[text=white] at (axis cs: 1,6) {-2.7};
        \node[text=white] at (axis cs: 2,6) {-0.2};
        \node[text=white] at (axis cs: 3,6) {-2.4};
        \node[text=white] at (axis cs: 4,6) {0.4};
        \node[text=white] at (axis cs: 5,6) {-2.0};
        \node[text=black] at (axis cs: 0,7) {-14.3};
        \node[text=white] at (axis cs: 1,7) {-1.2};
        \node[text=white] at (axis cs: 2,7) {2.9};
        \node[text=white] at (axis cs: 3,7) {-1.9};
        \node[text=white] at (axis cs: 4,7) {9.6};
        \node[text=white] at (axis cs: 5,7) {0.5};
        \node[text=white] at (axis cs: 0,8) {-7.7};
        \node[text=white] at (axis cs: 1,8) {-1.7};
        \node[text=white] at (axis cs: 2,8) {-0.1};
        \node[text=white] at (axis cs: 3,8) {-1.0};
        \node[text=white] at (axis cs: 4,8) {9.0};
        \node[text=white] at (axis cs: 5,8) {0.1};
        \node[text=black] at (axis cs: 0,9) {-17.5};
        \node[text=white] at (axis cs: 1,9) {-2.2};
        \node[text=white] at (axis cs: 2,9) {0.0};
        \node[text=white] at (axis cs: 3,9) {6.1};
        \node[text=white] at (axis cs: 4,9) {-1.9};
        \node[text=white] at (axis cs: 5,9) {-1.3};
        \node[text=white] at (axis cs: 0,10) {-7.6};
        \node[text=white] at (axis cs: 1,10) {-2.2};
        \node[text=white] at (axis cs: 2,10) {0.6};
        \node[text=white] at (axis cs: 3,10) {5.6};
        \node[text=white] at (axis cs: 4,10) {8.4};
        \node[text=white] at (axis cs: 5,10) {0.3};
        \node[text=black] at (axis cs: 0,11) {-43.2};
        \node[text=white] at (axis cs: 1,11) {-1.3};
        \node[text=white] at (axis cs: 2,11) {-0.6};
        \node[text=white] at (axis cs: 3,11) {0.3};
        \node[text=white] at (axis cs: 4,11) {6.3};
        \node[text=white] at (axis cs: 5,11) {0.0};
        \node[text=black] at (axis cs: 0,12) {-21.0};
        \node[text=white] at (axis cs: 1,12) {-1.4};
        \node[text=white] at (axis cs: 2,12) {0.0};
        \node[text=white] at (axis cs: 3,12) {4.6};
        \node[text=white] at (axis cs: 4,12) {7.0};
        \node[text=white] at (axis cs: 5,12) {1.7};
        \node[text=white] at (axis cs: 0,13) {-11.6};
        \node[text=white] at (axis cs: 1,13) {0.8};
        \node[text=white] at (axis cs: 2,13) {-5.3};
        \node[text=white] at (axis cs: 3,13) {3.8};
        \node[text=white] at (axis cs: 4,13) {-2.0};
        \node[text=white] at (axis cs: 5,13) {-2.3};
        \node[text=white] at (axis cs: 0,14) {-6.6};
        \node[text=white] at (axis cs: 1,14) {2.8};
        \node[text=white] at (axis cs: 2,14) {0.8};
        \node[text=white] at (axis cs: 3,14) {2.8};
        \node[text=white] at (axis cs: 4,14) {8.1};
        \node[text=white] at (axis cs: 5,14) {-0.7};
        \node[text=black] at (axis cs: 0,15) {-14.4};
        \node[text=white] at (axis cs: 1,15) {-2.4};
        \node[text=white] at (axis cs: 2,15) {1.9};
        \node[text=white] at (axis cs: 3,15) {1.0};
        \node[text=white] at (axis cs: 4,15) {0.0};
        \node[text=white] at (axis cs: 5,15) {-1.6};
        \node[text=black] at (axis cs: 0,16) {-13.0};
        \node[text=white] at (axis cs: 1,16) {-3.9};
        \node[text=white] at (axis cs: 2,16) {-2.5};
        \node[text=white] at (axis cs: 3,16) {1.1};
        \node[text=white] at (axis cs: 4,16) {10.5};
        \node[text=white] at (axis cs: 5,16) {0.5};
        \node[text=black] at (axis cs: 0,17) {-27.8};
        \node[text=white] at (axis cs: 1,17) {-3.9};
        \node[text=white] at (axis cs: 2,17) {0.5};
        \node[text=white] at (axis cs: 3,17) {-1.9};
        \node[text=white] at (axis cs: 4,17) {19.8};
        \node[text=white] at (axis cs: 5,17) {0.0};

        \fill[white] (-0.5,-0.5) rectangle (0.5,0.5);
        \fill[pattern=north west lines, pattern color=black] (-0.5,-0.5) rectangle (0.5,0.5);
        
        \fill[white] (0.5,0.5) rectangle (1.5,1.5);
        \fill[pattern=north west lines, pattern color=black] (0.5,0.5) rectangle (1.5,1.5);
        
        \fill[white] (1.5,1.5) rectangle (2.5,2.5);
        \fill[pattern=north west lines, pattern color=black] (1.5,1.5) rectangle (2.5,2.5);

        \fill[white] (2.5,2.5) rectangle (3.5,3.5);
        \fill[pattern=north west lines, pattern color=black] (2.5,2.5) rectangle (3.5,3.5);

        \fill[white] (3.5,14.5) rectangle (4.5,15.5);
        \fill[pattern=north west lines, pattern color=black] (3.5,14.5) rectangle (4.5,15.5);

        \fill[white] (4.5,16.5) rectangle (5.5,17.5);
        \fill[pattern=north west lines, pattern color=black] (4.5,16.5) rectangle (5.5,17.5);

    \end{axis}
\end{tikzpicture}
\caption{Regression task heat map of transfer performance of empathy source adapters on empathy target tasks. Values are calculated as the percent improvement of the transfer model over the baseline performance measured by Pearson $r$ correlation.}
\label{fig:heatmap_transfer_performances_regression}
\end{figure}

%% file: resources/similar_grid.tex
\begin{figure*}[t]
    \centering
    % \scriptsize
\begin{tikzpicture}
    \begin{axis}[enlargelimits=false,width=9cm,colorbar,colormap/Greys,
        xtick={0,1,2,3,4,5,6,7,8,9,10}, % {0,1,2,3,4,5,6}
        xticklabels={Condolence,Conv,EPT,EmpDial,EmpIntent,EmpQT,Empathy Hope,MI,News}, %  {CONV,Buchel,MI,ED,EPI}
        ytick={0,1,2,3,4,5,6,7,8,9,10},
        yticklabels={Condolence,Conv,EPT,EmpDial,EmpIntent,EmpQT,Empathy Hope,MI,News}, %{CONV,Buchel,MI,ED,EPI},
        xticklabel style={rotate=90},
        every node near coord/.append style={yshift=-0.25cm},
        ytick style={draw=none}, xtick style={draw=none},
    ]
    \addplot [matrix plot,point meta=explicit, ]
        coordinates {
            % (x,y) position and the [cell value]
            % (x0,y0) [v0] (x1,y0) [v1] ... (xn,y0)

            % (x0,y1) [v0] (x1,y1) [v1] ... (xn,y1)
            % IMPORTANT: make sure there is an empty line between each row
            (0,0) [1.00] (1,0) [0.89] (2,0) [0.91] (3,0) [0.82] (4,0) [0.81] (5,0) [0.76] (6,0) [0.62] (7,0) [0.95] (8,0) [0.83]

                (0,1) [0.89] (1,1) [1.00] (2,1) [0.89] (3,1) [0.73] (4,1) [0.76] (5,1) [0.74] (6,1) [0.61] (7,1) [0.91] (8,1) [0.88]

                (0,2) [0.91] (1,2) [0.89] (2,2) [1.00] (3,2) [0.81] (4,2) [0.83] (5,2) [0.80] (6,2) [0.60] (7,2) [0.95] (8,2) [0.83]

                (0,3) [0.82] (1,3) [0.73] (2,3) [0.81] (3,3) [1.00] (4,3) [0.86] (5,3) [0.83] (6,3) [0.56] (7,3) [0.81] (8,3) [0.77]

                (0,4) [0.81] (1,4) [0.76] (2,4) [0.83] (3,4) [0.86] (4,4) [1.00] (5,4) [0.99] (6,4) [0.50] (7,4) [0.82] (8,4) [0.68]

                (0,5) [0.76] (1,5) [0.74] (2,5) [0.80] (3,5) [0.83] (4,5) [0.99] (5,5) [1.00] (6,5) [0.47] (7,5) [0.78] (8,5) [0.65]

                (0,6) [0.62] (1,6) [0.61] (2,6) [0.60] (3,6) [0.56] (4,6) [0.50] (5,6) [0.47] (6,6) [1.00] (7,6) [0.59] (8,6) [0.62]

                (0,7) [0.95] (1,7) [0.91] (2,7) [0.95] (3,7) [0.81] (4,7) [0.82] (5,7) [0.78] (6,7) [0.59] (7,7) [1.00] (8,7) [0.85]

                (0,8) [0.83] (1,8) [0.88] (2,8) [0.83] (3,8) [0.77] (4,8) [0.68] (5,8) [0.65] (6,8) [0.62] (7,8) [0.85] (8,8) [1.00]
        };
    % white text if value >= .7ish, black if smaller
    % \node[text=white if value >= .7ish else black]
    % for each y, for each x:
    %   \node[text=color] at (axis cs: x,y) {v};
    \node[text=white] at (axis cs: 0,0) {1.00};
        \node[text=white] at (axis cs: 1,0) {0.89};
        \node[text=white] at (axis cs: 2,0) {0.91};
        \node[text=white] at (axis cs: 3,0) {0.82};
        \node[text=white] at (axis cs: 4,0) {0.81};
        \node[text=white] at (axis cs: 5,0) {0.76};
        \node[text=black] at (axis cs: 6,0) {0.62};
        \node[text=white] at (axis cs: 7,0) {0.95};
        \node[text=white] at (axis cs: 8,0) {0.83};
        \node[text=white] at (axis cs: 0,1) {0.89};
        \node[text=white] at (axis cs: 1,1) {1.00};
        \node[text=white] at (axis cs: 2,1) {0.89};
        \node[text=black] at (axis cs: 3,1) {0.73};
        \node[text=white] at (axis cs: 4,1) {0.76};
        \node[text=white] at (axis cs: 5,1) {0.74};
        \node[text=black] at (axis cs: 6,1) {0.61};
        \node[text=white] at (axis cs: 7,1) {0.91};
        \node[text=white] at (axis cs: 8,1) {0.88};
        \node[text=white] at (axis cs: 0,2) {0.91};
        \node[text=white] at (axis cs: 1,2) {0.89};
        \node[text=white] at (axis cs: 2,2) {1.00};
        \node[text=white] at (axis cs: 3,2) {0.81};
        \node[text=white] at (axis cs: 4,2) {0.83};
        \node[text=white] at (axis cs: 5,2) {0.80};
        \node[text=black] at (axis cs: 6,2) {0.60};
        \node[text=white] at (axis cs: 7,2) {0.95};
        \node[text=white] at (axis cs: 8,2) {0.83};
        \node[text=white] at (axis cs: 0,3) {0.82};
        \node[text=black] at (axis cs: 1,3) {0.73};
        \node[text=white] at (axis cs: 2,3) {0.81};
        \node[text=white] at (axis cs: 3,3) {1.00};
        \node[text=white] at (axis cs: 4,3) {0.86};
        \node[text=white] at (axis cs: 5,3) {0.83};
        \node[text=black] at (axis cs: 6,3) {0.56};
        \node[text=white] at (axis cs: 7,3) {0.81};
        \node[text=white] at (axis cs: 8,3) {0.77};
        \node[text=white] at (axis cs: 0,4) {0.81};
        \node[text=white] at (axis cs: 1,4) {0.76};
        \node[text=white] at (axis cs: 2,4) {0.83};
        \node[text=white] at (axis cs: 3,4) {0.86};
        \node[text=white] at (axis cs: 4,4) {1.00};
        \node[text=white] at (axis cs: 5,4) {0.99};
        \node[text=black] at (axis cs: 6,4) {0.50};
        \node[text=white] at (axis cs: 7,4) {0.82};
        \node[text=black] at (axis cs: 8,4) {0.68};
        \node[text=white] at (axis cs: 0,5) {0.76};
        \node[text=white] at (axis cs: 1,5) {0.74};
        \node[text=white] at (axis cs: 2,5) {0.80};
        \node[text=white] at (axis cs: 3,5) {0.83};
        \node[text=white] at (axis cs: 4,5) {0.99};
        \node[text=white] at (axis cs: 5,5) {1.00};
        \node[text=black] at (axis cs: 6,5) {0.47};
        \node[text=white] at (axis cs: 7,5) {0.78};
        \node[text=black] at (axis cs: 8,5) {0.65};
        \node[text=black] at (axis cs: 0,6) {0.62};
        \node[text=black] at (axis cs: 1,6) {0.61};
        \node[text=black] at (axis cs: 2,6) {0.60};
        \node[text=black] at (axis cs: 3,6) {0.56};
        \node[text=black] at (axis cs: 4,6) {0.50};
        \node[text=black] at (axis cs: 5,6) {0.47};
        \node[text=white] at (axis cs: 6,6) {1.00};
        \node[text=black] at (axis cs: 7,6) {0.59};
        \node[text=black] at (axis cs: 8,6) {0.62};
        \node[text=white] at (axis cs: 0,7) {0.95};
        \node[text=white] at (axis cs: 1,7) {0.91};
        \node[text=white] at (axis cs: 2,7) {0.95};
        \node[text=white] at (axis cs: 3,7) {0.81};
        \node[text=white] at (axis cs: 4,7) {0.82};
        \node[text=white] at (axis cs: 5,7) {0.78};
        \node[text=black] at (axis cs: 6,7) {0.59};
        \node[text=white] at (axis cs: 7,7) {1.00};
        \node[text=white] at (axis cs: 8,7) {0.85};
        \node[text=white] at (axis cs: 0,8) {0.83};
        \node[text=white] at (axis cs: 1,8) {0.88};
        \node[text=white] at (axis cs: 2,8) {0.83};
        \node[text=white] at (axis cs: 3,8) {0.77};
        \node[text=black] at (axis cs: 4,8) {0.68};
        \node[text=black] at (axis cs: 5,8) {0.65};
        \node[text=black] at (axis cs: 6,8) {0.62};
        \node[text=white] at (axis cs: 7,8) {0.85};
        \node[text=white] at (axis cs: 8,8) {1.00};
    \end{axis}
\end{tikzpicture}
\caption{Heat map of dataset cosine similarity. Calculated with SBERT embeddings averaged over all dataset instances.}
    \label{fig:heatmap_sbert}
\end{figure*}

%% file: resources/task_emb_sim.tex
\begin{tikzpicture}
    \begin{axis}[enlargelimits=false,width=12cm,colorbar,colormap/Greys,
        xtick={0,1,2,3,4,5,6,7,8,9,10,11,12,13,14,15,16,17,18,19}, % {0,1,2,3,4,5,6}
        xticklabels={Condolence,Conv EmoInt,Conv EmoPol,Conv Empathy,EmpDial EI.,EmpDial Emo,EmpDial Role,Empathy Hope,Epitome ER,Epitome EX,Epitome IP,MI Adherent,MI Behavior,News Dis,News Emotion,News Empathy,EmpDial QAct.,EmpDial QInt.}, %  {CONV,Buchel,MI,ED,EPI}
        ytick={0,1,2,3,4,5,6,7,8,9,10,11,12,13,14,15,16,17,18,19},
        yticklabels={Condolence,Conv EmoInt,Conv EmoPol,Conv Empathy,EmpDial EI.,EmpDial Emo,EmpDial Role,Empathy Hope,Epitome ER,Epitome EX,Epitome IP,MI Adherent,MI Behavior,News Dis,News Emotion,News Empathy,EmpDial QAct.,EmpDial QInt.}, %{CONV,Buchel,MI,ED,EPI},
        xticklabel style={rotate=90},
        every node near coord/.append style={yshift=-0.25cm},
        ytick style={draw=none}, xtick style={draw=none},
    ]
    \addplot [matrix plot,point meta=explicit, ]
        coordinates {
            % (x,y) position and the [cell value]
            % (x0,y0) [v0] (x1,y0) [v1] ... (xn,y0)

            % (x0,y1) [v0] (x1,y1) [v1] ... (xn,y1)
            % IMPORTANT: make sure there is an empty line between each row
            (0,0) [0.82] (1,0) [0.70] (2,0) [0.72] (3,0) [0.70] (4,0) [0.70] (5,0) [0.70] (6,0) [0.70] (7,0) [0.71] (8,0) [0.73] (9,0) [0.71] (10,0) [0.69] (11,0) [0.70] (12,0) [0.70] (13,0) [0.70] (14,0) [0.69] (15,0) [0.70] (16,0) [0.71] (17,0) [0.70]

                (0,1) [0.71] (1,1) [0.82] (2,1) [0.72] (3,1) [0.71] (4,1) [0.70] (5,1) [0.71] (6,1) [0.70] (7,1) [0.70] (8,1) [0.69] (9,1) [0.70] (10,1) [0.71] (11,1) [0.70] (12,1) [0.69] (13,1) [0.70] (14,1) [0.71] (15,1) [0.71] (16,1) [0.71] (17,1) [0.70]

                (0,2) [0.70] (1,2) [0.70] (2,2) [0.82] (3,2) [0.71] (4,2) [0.70] (5,2) [0.69] (6,2) [0.69] (7,2) [0.69] (8,2) [0.69] (9,2) [0.69] (10,2) [0.69] (11,2) [0.69] (12,2) [0.68] (13,2) [0.69] (14,2) [0.67] (15,2) [0.69] (16,2) [0.69] (17,2) [0.70]

                (0,3) [0.70] (1,3) [0.70] (2,3) [0.72] (3,3) [0.82] (4,3) [0.70] (5,3) [0.69] (6,3) [0.70] (7,3) [0.69] (8,3) [0.70] (9,3) [0.69] (10,3) [0.71] (11,3) [0.69] (12,3) [0.68] (13,3) [0.71] (14,3) [0.69] (15,3) [0.69] (16,3) [0.70] (17,3) [0.71]

                (0,4) [0.73] (1,4) [0.73] (2,4) [0.75] (3,4) [0.74] (4,4) [0.82] (5,4) [0.74] (6,4) [0.75] (7,4) [0.72] (8,4) [0.73] (9,4) [0.74] (10,4) [0.75] (11,4) [0.74] (12,4) [0.74] (13,4) [0.73] (14,4) [0.74] (15,4) [0.73] (16,4) [0.73] (17,4) [0.74]

                (0,5) [0.75] (1,5) [0.76] (2,5) [0.75] (3,5) [0.74] (4,5) [0.75] (5,5) [0.82] (6,5) [0.73] (7,5) [0.76] (8,5) [0.76] (9,5) [0.74] (10,5) [0.74] (11,5) [0.76] (12,5) [0.75] (13,5) [0.75] (14,5) [0.76] (15,5) [0.75] (16,5) [0.75] (17,5) [0.75]

                (0,6) [0.71] (1,6) [0.69] (2,6) [0.70] (3,6) [0.71] (4,6) [0.73] (5,6) [0.69] (6,6) [0.82] (7,6) [0.70] (8,6) [0.71] (9,6) [0.73] (10,6) [0.72] (11,6) [0.72] (12,6) [0.70] (13,6) [0.70] (14,6) [0.70] (15,6) [0.71] (16,6) [0.71] (17,6) [0.71]

                (0,7) [0.71] (1,7) [0.71] (2,7) [0.70] (3,7) [0.70] (4,7) [0.71] (5,7) [0.73] (6,7) [0.70] (7,7) [0.82] (8,7) [0.73] (9,7) [0.71] (10,7) [0.71] (11,7) [0.70] (12,7) [0.71] (13,7) [0.71] (14,7) [0.74] (15,7) [0.72] (16,7) [0.72] (17,7) [0.73]

                (0,8) [0.75] (1,8) [0.72] (2,8) [0.73] (3,8) [0.73] (4,8) [0.72] (5,8) [0.74] (6,8) [0.73] (7,8) [0.74] (8,8) [0.82] (9,8) [0.72] (10,8) [0.73] (11,8) [0.73] (12,8) [0.73] (13,8) [0.73] (14,8) [0.74] (15,8) [0.72] (16,8) [0.73] (17,8) [0.73]

                (0,9) [0.72] (1,9) [0.72] (2,9) [0.72] (3,9) [0.71] (4,9) [0.73] (5,9) [0.71] (6,9) [0.74] (7,9) [0.71] (8,9) [0.71] (9,9) [0.82] (10,9) [0.72] (11,9) [0.73] (12,9) [0.72] (13,9) [0.72] (14,9) [0.73] (15,9) [0.72] (16,9) [0.72] (17,9) [0.73]

                (0,10) [0.71] (1,10) [0.73] (2,10) [0.72] (3,10) [0.73] (4,10) [0.74] (5,10) [0.72] (6,10) [0.74] (7,10) [0.72] (8,10) [0.72] (9,10) [0.72] (10,10) [0.82] (11,10) [0.72] (12,10) [0.74] (13,10) [0.72] (14,10) [0.74] (15,10) [0.73] (16,10) [0.74] (17,10) [0.73]

                (0,11) [0.70] (1,11) [0.69] (2,11) [0.69] (3,11) [0.69] (4,11) [0.69] (5,11) [0.70] (6,11) [0.70] (7,11) [0.68] (8,11) [0.70] (9,11) [0.69] (10,11) [0.69] (11,11) [0.82] (12,11) [0.68] (13,11) [0.70] (14,11) [0.69] (15,11) [0.68] (16,11) [0.69] (17,11) [0.69]

                (0,12) [0.70] (1,12) [0.69] (2,12) [0.69] (3,12) [0.69] (4,12) [0.69] (5,12) [0.69] (6,12) [0.68] (7,12) [0.70] (8,12) [0.69] (9,12) [0.69] (10,12) [0.70] (11,12) [0.69] (12,12) [0.82] (13,12) [0.69] (14,12) [0.70] (15,12) [0.69] (16,12) [0.69] (17,12) [0.69]

                (0,13) [0.70] (1,13) [0.71] (2,13) [0.69] (3,13) [0.71] (4,13) [0.70] (5,13) [0.71] (6,13) [0.69] (7,13) [0.70] (8,13) [0.71] (9,13) [0.70] (10,13) [0.69] (11,13) [0.71] (12,13) [0.70] (13,13) [0.82] (14,13) [0.70] (15,13) [0.73] (16,13) [0.71] (17,13) [0.70]

                (0,14) [0.73] (1,14) [0.75] (2,14) [0.72] (3,14) [0.74] (4,14) [0.74] (5,14) [0.76] (6,14) [0.73] (7,14) [0.77] (8,14) [0.75] (9,14) [0.75] (10,14) [0.75] (11,14) [0.74] (12,14) [0.75] (13,14) [0.74] (14,14) [0.82] (15,14) [0.75] (16,14) [0.75] (17,14) [0.74]

                (0,15) [0.70] (1,15) [0.70] (2,15) [0.70] (3,15) [0.69] (4,15) [0.69] (5,15) [0.70] (6,15) [0.71] (7,15) [0.71] (8,15) [0.69] (9,15) [0.70] (10,15) [0.70] (11,15) [0.69] (12,15) [0.70] (13,15) [0.72] (14,15) [0.70] (15,15) [0.82] (16,15) [0.70] (17,15) [0.70]

                (0,16) [0.77] (1,16) [0.75] (2,16) [0.75] (3,16) [0.76] (4,16) [0.75] (5,16) [0.76] (6,16) [0.76] (7,16) [0.75] (8,16) [0.76] (9,16) [0.76] (10,16) [0.76] (11,16) [0.76] (12,16) [0.77] (13,16) [0.76] (14,16) [0.76] (15,16) [0.75] (16,16) [0.82] (17,16) [0.77]

                (0,17) [0.74] (1,17) [0.75] (2,17) [0.75] (3,17) [0.75] (4,17) [0.75] (5,17) [0.76] (6,17) [0.74] (7,17) [0.75] (8,17) [0.74] (9,17) [0.75] (10,17) [0.75] (11,17) [0.73] (12,17) [0.75] (13,17) [0.73] (14,17) [0.74] (15,17) [0.74] (16,17) [0.76] (17,17) [0.82]
        };
    % white text if value >= .7ish, black if smaller
    % \node[text=white if value >= .7ish else black]
    % for each y, for each x:
    %   \node[text=color] at (axis cs: x,y) {v};
    \node[text=white] at (axis cs: 0,0) {0.82};
        \node[text=black] at (axis cs: 1,0) {0.70};
        \node[text=black] at (axis cs: 2,0) {0.72};
        \node[text=black] at (axis cs: 3,0) {0.70};
        \node[text=black] at (axis cs: 4,0) {0.70};
        \node[text=black] at (axis cs: 5,0) {0.70};
        \node[text=black] at (axis cs: 6,0) {0.70};
        \node[text=black] at (axis cs: 7,0) {0.71};
        \node[text=black] at (axis cs: 8,0) {0.73};
        \node[text=black] at (axis cs: 9,0) {0.71};
        \node[text=black] at (axis cs: 10,0) {0.69};
        \node[text=black] at (axis cs: 11,0) {0.70};
        \node[text=black] at (axis cs: 12,0) {0.70};
        \node[text=black] at (axis cs: 13,0) {0.70};
        \node[text=black] at (axis cs: 14,0) {0.69};
        \node[text=black] at (axis cs: 15,0) {0.70};
        \node[text=black] at (axis cs: 16,0) {0.71};
        \node[text=black] at (axis cs: 17,0) {0.70};
        \node[text=black] at (axis cs: 0,1) {0.71};
        \node[text=white] at (axis cs: 1,1) {0.82};
        \node[text=black] at (axis cs: 2,1) {0.72};
        \node[text=black] at (axis cs: 3,1) {0.71};
        \node[text=black] at (axis cs: 4,1) {0.70};
        \node[text=black] at (axis cs: 5,1) {0.71};
        \node[text=black] at (axis cs: 6,1) {0.70};
        \node[text=black] at (axis cs: 7,1) {0.70};
        \node[text=black] at (axis cs: 8,1) {0.69};
        \node[text=black] at (axis cs: 9,1) {0.70};
        \node[text=black] at (axis cs: 10,1) {0.71};
        \node[text=black] at (axis cs: 11,1) {0.70};
        \node[text=black] at (axis cs: 12,1) {0.69};
        \node[text=black] at (axis cs: 13,1) {0.70};
        \node[text=black] at (axis cs: 14,1) {0.71};
        \node[text=black] at (axis cs: 15,1) {0.71};
        \node[text=black] at (axis cs: 16,1) {0.71};
        \node[text=black] at (axis cs: 17,1) {0.70};
        \node[text=black] at (axis cs: 0,2) {0.70};
        \node[text=black] at (axis cs: 1,2) {0.70};
        \node[text=white] at (axis cs: 2,2) {0.82};
        \node[text=black] at (axis cs: 3,2) {0.71};
        \node[text=black] at (axis cs: 4,2) {0.70};
        \node[text=black] at (axis cs: 5,2) {0.69};
        \node[text=black] at (axis cs: 6,2) {0.69};
        \node[text=black] at (axis cs: 7,2) {0.69};
        \node[text=black] at (axis cs: 8,2) {0.69};
        \node[text=black] at (axis cs: 9,2) {0.69};
        \node[text=black] at (axis cs: 10,2) {0.69};
        \node[text=black] at (axis cs: 11,2) {0.69};
        \node[text=black] at (axis cs: 12,2) {0.68};
        \node[text=black] at (axis cs: 13,2) {0.69};
        \node[text=black] at (axis cs: 14,2) {0.67};
        \node[text=black] at (axis cs: 15,2) {0.69};
        \node[text=black] at (axis cs: 16,2) {0.69};
        \node[text=black] at (axis cs: 17,2) {0.70};
        \node[text=black] at (axis cs: 0,3) {0.70};
        \node[text=black] at (axis cs: 1,3) {0.70};
        \node[text=black] at (axis cs: 2,3) {0.72};
        \node[text=white] at (axis cs: 3,3) {0.82};
        \node[text=black] at (axis cs: 4,3) {0.70};
        \node[text=black] at (axis cs: 5,3) {0.69};
        \node[text=black] at (axis cs: 6,3) {0.70};
        \node[text=black] at (axis cs: 7,3) {0.69};
        \node[text=black] at (axis cs: 8,3) {0.70};
        \node[text=black] at (axis cs: 9,3) {0.69};
        \node[text=black] at (axis cs: 10,3) {0.71};
        \node[text=black] at (axis cs: 11,3) {0.69};
        \node[text=black] at (axis cs: 12,3) {0.68};
        \node[text=black] at (axis cs: 13,3) {0.71};
        \node[text=black] at (axis cs: 14,3) {0.69};
        \node[text=black] at (axis cs: 15,3) {0.69};
        \node[text=black] at (axis cs: 16,3) {0.70};
        \node[text=black] at (axis cs: 17,3) {0.71};
        \node[text=black] at (axis cs: 0,4) {0.73};
        \node[text=black] at (axis cs: 1,4) {0.73};
        \node[text=white] at (axis cs: 2,4) {0.75};
        \node[text=black] at (axis cs: 3,4) {0.74};
        \node[text=white] at (axis cs: 4,4) {0.82};
        \node[text=black] at (axis cs: 5,4) {0.74};
        \node[text=white] at (axis cs: 6,4) {0.75};
        \node[text=black] at (axis cs: 7,4) {0.72};
        \node[text=black] at (axis cs: 8,4) {0.73};
        \node[text=black] at (axis cs: 9,4) {0.74};
        \node[text=white] at (axis cs: 10,4) {0.75};
        \node[text=black] at (axis cs: 11,4) {0.74};
        \node[text=black] at (axis cs: 12,4) {0.74};
        \node[text=black] at (axis cs: 13,4) {0.73};
        \node[text=black] at (axis cs: 14,4) {0.74};
        \node[text=black] at (axis cs: 15,4) {0.73};
        \node[text=black] at (axis cs: 16,4) {0.73};
        \node[text=black] at (axis cs: 17,4) {0.74};
        \node[text=white] at (axis cs: 0,5) {0.75};
        \node[text=white] at (axis cs: 1,5) {0.76};
        \node[text=black] at (axis cs: 2,5) {0.75};
        \node[text=black] at (axis cs: 3,5) {0.74};
        \node[text=white] at (axis cs: 4,5) {0.75};
        \node[text=white] at (axis cs: 5,5) {0.82};
        \node[text=black] at (axis cs: 6,5) {0.73};
        \node[text=white] at (axis cs: 7,5) {0.76};
        \node[text=white] at (axis cs: 8,5) {0.76};
        \node[text=black] at (axis cs: 9,5) {0.74};
        \node[text=black] at (axis cs: 10,5) {0.74};
        \node[text=white] at (axis cs: 11,5) {0.76};
        \node[text=white] at (axis cs: 12,5) {0.75};
        \node[text=white] at (axis cs: 13,5) {0.75};
        \node[text=white] at (axis cs: 14,5) {0.76};
        \node[text=white] at (axis cs: 15,5) {0.75};
        \node[text=white] at (axis cs: 16,5) {0.75};
        \node[text=white] at (axis cs: 17,5) {0.75};
        \node[text=black] at (axis cs: 0,6) {0.71};
        \node[text=black] at (axis cs: 1,6) {0.69};
        \node[text=black] at (axis cs: 2,6) {0.70};
        \node[text=black] at (axis cs: 3,6) {0.71};
        \node[text=black] at (axis cs: 4,6) {0.73};
        \node[text=black] at (axis cs: 5,6) {0.69};
        \node[text=white] at (axis cs: 6,6) {0.82};
        \node[text=black] at (axis cs: 7,6) {0.70};
        \node[text=black] at (axis cs: 8,6) {0.71};
        \node[text=black] at (axis cs: 9,6) {0.73};
        \node[text=black] at (axis cs: 10,6) {0.72};
        \node[text=black] at (axis cs: 11,6) {0.72};
        \node[text=black] at (axis cs: 12,6) {0.70};
        \node[text=black] at (axis cs: 13,6) {0.70};
        \node[text=black] at (axis cs: 14,6) {0.70};
        \node[text=black] at (axis cs: 15,6) {0.71};
        \node[text=black] at (axis cs: 16,6) {0.71};
        \node[text=black] at (axis cs: 17,6) {0.71};
        \node[text=black] at (axis cs: 0,7) {0.71};
        \node[text=black] at (axis cs: 1,7) {0.71};
        \node[text=black] at (axis cs: 2,7) {0.70};
        \node[text=black] at (axis cs: 3,7) {0.70};
        \node[text=black] at (axis cs: 4,7) {0.71};
        \node[text=black] at (axis cs: 5,7) {0.73};
        \node[text=black] at (axis cs: 6,7) {0.70};
        \node[text=white] at (axis cs: 7,7) {0.82};
        \node[text=black] at (axis cs: 8,7) {0.73};
        \node[text=black] at (axis cs: 9,7) {0.71};
        \node[text=black] at (axis cs: 10,7) {0.71};
        \node[text=black] at (axis cs: 11,7) {0.70};
        \node[text=black] at (axis cs: 12,7) {0.71};
        \node[text=black] at (axis cs: 13,7) {0.71};
        \node[text=black] at (axis cs: 14,7) {0.74};
        \node[text=black] at (axis cs: 15,7) {0.72};
        \node[text=black] at (axis cs: 16,7) {0.72};
        \node[text=black] at (axis cs: 17,7) {0.73};
        \node[text=white] at (axis cs: 0,8) {0.75};
        \node[text=black] at (axis cs: 1,8) {0.72};
        \node[text=black] at (axis cs: 2,8) {0.73};
        \node[text=black] at (axis cs: 3,8) {0.73};
        \node[text=black] at (axis cs: 4,8) {0.72};
        \node[text=black] at (axis cs: 5,8) {0.74};
        \node[text=black] at (axis cs: 6,8) {0.73};
        \node[text=black] at (axis cs: 7,8) {0.74};
        \node[text=white] at (axis cs: 8,8) {0.82};
        \node[text=black] at (axis cs: 9,8) {0.72};
        \node[text=black] at (axis cs: 10,8) {0.73};
        \node[text=black] at (axis cs: 11,8) {0.73};
        \node[text=black] at (axis cs: 12,8) {0.73};
        \node[text=black] at (axis cs: 13,8) {0.73};
        \node[text=black] at (axis cs: 14,8) {0.74};
        \node[text=black] at (axis cs: 15,8) {0.72};
        \node[text=black] at (axis cs: 16,8) {0.73};
        \node[text=black] at (axis cs: 17,8) {0.73};
        \node[text=black] at (axis cs: 0,9) {0.72};
        \node[text=black] at (axis cs: 1,9) {0.72};
        \node[text=black] at (axis cs: 2,9) {0.72};
        \node[text=black] at (axis cs: 3,9) {0.71};
        \node[text=black] at (axis cs: 4,9) {0.73};
        \node[text=black] at (axis cs: 5,9) {0.71};
        \node[text=black] at (axis cs: 6,9) {0.74};
        \node[text=black] at (axis cs: 7,9) {0.71};
        \node[text=black] at (axis cs: 8,9) {0.71};
        \node[text=white] at (axis cs: 9,9) {0.82};
        \node[text=black] at (axis cs: 10,9) {0.72};
        \node[text=black] at (axis cs: 11,9) {0.73};
        \node[text=black] at (axis cs: 12,9) {0.72};
        \node[text=black] at (axis cs: 13,9) {0.72};
        \node[text=black] at (axis cs: 14,9) {0.73};
        \node[text=black] at (axis cs: 15,9) {0.72};
        \node[text=black] at (axis cs: 16,9) {0.72};
        \node[text=black] at (axis cs: 17,9) {0.73};
        \node[text=black] at (axis cs: 0,10) {0.71};
        \node[text=black] at (axis cs: 1,10) {0.73};
        \node[text=black] at (axis cs: 2,10) {0.72};
        \node[text=black] at (axis cs: 3,10) {0.73};
        \node[text=black] at (axis cs: 4,10) {0.74};
        \node[text=black] at (axis cs: 5,10) {0.72};
        \node[text=black] at (axis cs: 6,10) {0.74};
        \node[text=black] at (axis cs: 7,10) {0.72};
        \node[text=black] at (axis cs: 8,10) {0.72};
        \node[text=black] at (axis cs: 9,10) {0.72};
        \node[text=white] at (axis cs: 10,10) {0.82};
        \node[text=black] at (axis cs: 11,10) {0.72};
        \node[text=black] at (axis cs: 12,10) {0.74};
        \node[text=black] at (axis cs: 13,10) {0.72};
        \node[text=black] at (axis cs: 14,10) {0.74};
        \node[text=black] at (axis cs: 15,10) {0.73};
        \node[text=black] at (axis cs: 16,10) {0.74};
        \node[text=black] at (axis cs: 17,10) {0.73};
        \node[text=black] at (axis cs: 0,11) {0.70};
        \node[text=black] at (axis cs: 1,11) {0.69};
        \node[text=black] at (axis cs: 2,11) {0.69};
        \node[text=black] at (axis cs: 3,11) {0.69};
        \node[text=black] at (axis cs: 4,11) {0.69};
        \node[text=black] at (axis cs: 5,11) {0.70};
        \node[text=black] at (axis cs: 6,11) {0.70};
        \node[text=black] at (axis cs: 7,11) {0.68};
        \node[text=black] at (axis cs: 8,11) {0.70};
        \node[text=black] at (axis cs: 9,11) {0.69};
        \node[text=black] at (axis cs: 10,11) {0.69};
        \node[text=white] at (axis cs: 11,11) {0.82};
        \node[text=black] at (axis cs: 12,11) {0.68};
        \node[text=black] at (axis cs: 13,11) {0.70};
        \node[text=black] at (axis cs: 14,11) {0.69};
        \node[text=black] at (axis cs: 15,11) {0.68};
        \node[text=black] at (axis cs: 16,11) {0.69};
        \node[text=black] at (axis cs: 17,11) {0.69};
        \node[text=black] at (axis cs: 0,12) {0.70};
        \node[text=black] at (axis cs: 1,12) {0.69};
        \node[text=black] at (axis cs: 2,12) {0.69};
        \node[text=black] at (axis cs: 3,12) {0.69};
        \node[text=black] at (axis cs: 4,12) {0.69};
        \node[text=black] at (axis cs: 5,12) {0.69};
        \node[text=black] at (axis cs: 6,12) {0.68};
        \node[text=black] at (axis cs: 7,12) {0.70};
        \node[text=black] at (axis cs: 8,12) {0.69};
        \node[text=black] at (axis cs: 9,12) {0.69};
        \node[text=black] at (axis cs: 10,12) {0.70};
        \node[text=black] at (axis cs: 11,12) {0.69};
        \node[text=white] at (axis cs: 12,12) {0.82};
        \node[text=black] at (axis cs: 13,12) {0.69};
        \node[text=black] at (axis cs: 14,12) {0.70};
        \node[text=black] at (axis cs: 15,12) {0.69};
        \node[text=black] at (axis cs: 16,12) {0.69};
        \node[text=black] at (axis cs: 17,12) {0.69};
        \node[text=black] at (axis cs: 0,13) {0.70};
        \node[text=black] at (axis cs: 1,13) {0.71};
        \node[text=black] at (axis cs: 2,13) {0.69};
        \node[text=black] at (axis cs: 3,13) {0.71};
        \node[text=black] at (axis cs: 4,13) {0.70};
        \node[text=black] at (axis cs: 5,13) {0.71};
        \node[text=black] at (axis cs: 6,13) {0.69};
        \node[text=black] at (axis cs: 7,13) {0.70};
        \node[text=black] at (axis cs: 8,13) {0.71};
        \node[text=black] at (axis cs: 9,13) {0.70};
        \node[text=black] at (axis cs: 10,13) {0.69};
        \node[text=black] at (axis cs: 11,13) {0.71};
        \node[text=black] at (axis cs: 12,13) {0.70};
        \node[text=white] at (axis cs: 13,13) {0.82};
        \node[text=black] at (axis cs: 14,13) {0.70};
        \node[text=black] at (axis cs: 15,13) {0.73};
        \node[text=black] at (axis cs: 16,13) {0.71};
        \node[text=black] at (axis cs: 17,13) {0.70};
        \node[text=black] at (axis cs: 0,14) {0.73};
        \node[text=white] at (axis cs: 1,14) {0.75};
        \node[text=black] at (axis cs: 2,14) {0.72};
        \node[text=black] at (axis cs: 3,14) {0.74};
        \node[text=black] at (axis cs: 4,14) {0.74};
        \node[text=white] at (axis cs: 5,14) {0.76};
        \node[text=black] at (axis cs: 6,14) {0.73};
        \node[text=white] at (axis cs: 7,14) {0.77};
        \node[text=white] at (axis cs: 8,14) {0.75};
        \node[text=black] at (axis cs: 9,14) {0.75};
        \node[text=white] at (axis cs: 10,14) {0.75};
        \node[text=black] at (axis cs: 11,14) {0.74};
        \node[text=white] at (axis cs: 12,14) {0.75};
        \node[text=black] at (axis cs: 13,14) {0.74};
        \node[text=white] at (axis cs: 14,14) {0.82};
        \node[text=white] at (axis cs: 15,14) {0.75};
        \node[text=white] at (axis cs: 16,14) {0.75};
        \node[text=black] at (axis cs: 17,14) {0.74};
        \node[text=black] at (axis cs: 0,15) {0.70};
        \node[text=black] at (axis cs: 1,15) {0.70};
        \node[text=black] at (axis cs: 2,15) {0.70};
        \node[text=black] at (axis cs: 3,15) {0.69};
        \node[text=black] at (axis cs: 4,15) {0.69};
        \node[text=black] at (axis cs: 5,15) {0.70};
        \node[text=black] at (axis cs: 6,15) {0.71};
        \node[text=black] at (axis cs: 7,15) {0.71};
        \node[text=black] at (axis cs: 8,15) {0.69};
        \node[text=black] at (axis cs: 9,15) {0.70};
        \node[text=black] at (axis cs: 10,15) {0.70};
        \node[text=black] at (axis cs: 11,15) {0.69};
        \node[text=black] at (axis cs: 12,15) {0.70};
        \node[text=black] at (axis cs: 13,15) {0.72};
        \node[text=black] at (axis cs: 14,15) {0.70};
        \node[text=white] at (axis cs: 15,15) {0.82};
        \node[text=black] at (axis cs: 16,15) {0.70};
        \node[text=black] at (axis cs: 17,15) {0.70};
        \node[text=white] at (axis cs: 0,16) {0.77};
        \node[text=white] at (axis cs: 1,16) {0.75};
        \node[text=black] at (axis cs: 2,16) {0.75};
        \node[text=white] at (axis cs: 3,16) {0.76};
        \node[text=white] at (axis cs: 4,16) {0.75};
        \node[text=white] at (axis cs: 5,16) {0.76};
        \node[text=white] at (axis cs: 6,16) {0.76};
        \node[text=white] at (axis cs: 7,16) {0.75};
        \node[text=white] at (axis cs: 8,16) {0.76};
        \node[text=white] at (axis cs: 9,16) {0.76};
        \node[text=white] at (axis cs: 10,16) {0.76};
        \node[text=white] at (axis cs: 11,16) {0.76};
        \node[text=white] at (axis cs: 12,16) {0.77};
        \node[text=white] at (axis cs: 13,16) {0.76};
        \node[text=white] at (axis cs: 14,16) {0.76};
        \node[text=white] at (axis cs: 15,16) {0.75};
        \node[text=white] at (axis cs: 16,16) {0.82};
        \node[text=white] at (axis cs: 17,16) {0.77};
        \node[text=black] at (axis cs: 0,17) {0.74};
        \node[text=white] at (axis cs: 1,17) {0.75};
        \node[text=white] at (axis cs: 2,17) {0.75};
        \node[text=white] at (axis cs: 3,17) {0.75};
        \node[text=white] at (axis cs: 4,17) {0.75};
        \node[text=white] at (axis cs: 5,17) {0.76};
        \node[text=black] at (axis cs: 6,17) {0.74};
        \node[text=white] at (axis cs: 7,17) {0.75};
        \node[text=black] at (axis cs: 8,17) {0.74};
        \node[text=white] at (axis cs: 9,17) {0.75};
        \node[text=white] at (axis cs: 10,17) {0.75};
        \node[text=black] at (axis cs: 11,17) {0.73};
        \node[text=white] at (axis cs: 12,17) {0.75};
        \node[text=black] at (axis cs: 13,17) {0.73};
        \node[text=black] at (axis cs: 14,17) {0.74};
        \node[text=black] at (axis cs: 15,17) {0.74};
        \node[text=white] at (axis cs: 16,17) {0.76};
        \node[text=white] at (axis cs: 17,17) {0.82};
    \end{axis}
\end{tikzpicture}

%% file: resources/label_distribution_tables.tex
\begin{table*}[ht]
    \centering
    \begin{tabular}{c|c}
\toprule
      \begin{tabular}{lrrrr}
\multicolumn{5}{c}{\textbf{Condolence}} \\
        & All    & Train   & Val   & Test   \\
\midrule
 Split   & 100.0 & 79.8   & 10.2 & 10.1  \\
 Min    & 1.00   & 1.00    & 1.00  & 1.00   \\
 Median & 1.50   & 1.50    & 1.50  & 1.50   \\
 Max    & 5.00   & 5.00    & 5.00  & 5.00   \\
 Mean   & 1.54   & 1.53    & 1.57  & 1.56   \\
 std    & 0.76   & 0.75    & 0.83  & 0.82   \\
\end{tabular}   &  \begin{tabular}{lrrrr}
\multicolumn{5}{c}{\textbf{Conv EmoInt}} \\
        & All    & Train   & Val   & Test   \\
\midrule
 Split   & 100.0 & 80.1   & 10.3 & 9.6   \\
 Min    & 0.00   & 0.00    & 0.00  & 0.00   \\
 Median & 2.00   & 2.33    & 2.00  & 2.00   \\
 Max    & 5.00   & 5.00    & 5.00  & 4.67   \\
 Mean   & 2.22   & 2.24    & 2.19  & 2.09   \\
 std    & 0.82   & 0.82    & 0.88  & 0.79   \\
\end{tabular} \\
 \begin{tabular}{lrrrr}
\toprule
\multicolumn{5}{c}{\textbf{Conv EmoPol}} \\
        & All    & Train   & Val   & Test   \\
\midrule
 Split   & 100.0 & 80.1   & 10.3 & 9.6   \\
 Min    & 0.00   & 0.00    & 0.00  & 0.00   \\
 Median & 1.33   & 1.33    & 1.33  & 1.33   \\
 Max    & 2.00   & 2.00    & 2.00  & 2.00   \\
 Mean   & 1.29   & 1.29    & 1.28  & 1.28   \\
 std    & 0.61   & 0.61    & 0.61  & 0.60   \\
\end{tabular}        & \begin{tabular}{lrrrr}
\toprule
\multicolumn{5}{c}{\textbf{Conv Empathy}} \\
        & All    & Train   & Val   & Test   \\
\midrule
 Split   & 100.0 & 80.1   & 10.3 & 9.6   \\
 Min    & 0.00   & 0.00    & 0.00  & 0.00   \\
 Median & 2.00   & 2.00    & 2.00  & 2.00   \\
 Max    & 5.00   & 5.00    & 4.67  & 4.33   \\
 Mean   & 2.09   & 2.08    & 2.17  & 2.02   \\
 std    & 0.92   & 0.92    & 0.92  & 0.91   \\
\end{tabular} \\
\begin{tabular}{lrrrr}
\toprule
\multicolumn{5}{c}{\textbf{News Distress}} \\
        & All    & Train   & Val   & Test   \\
\midrule
 Split   & 100.0 & 29.6   & 37.0 & 33.3  \\
 Min    & 1.00   & 1.00    & 1.00  & 1.00   \\
 Median & 3.75   & 3.62    & 3.75  & 3.69   \\
 Max    & 7.00   & 7.00    & 7.00  & 7.00   \\
 Mean   & 3.76   & 3.74    & 3.79  & 3.74   \\
 std    & 2.00   & 2.01    & 2.00  & 2.00   \\
\end{tabular} & \begin{tabular}{lrrrr}
\toprule
\multicolumn{5}{c}{\textbf{News Empathy}} \\
        & All    & Train   & Val   & Test   \\
\midrule
 Split   & 100.0 & 29.6   & 37.0 & 33.3  \\
 Min    & 1.00   & 1.00    & 1.00  & 1.00   \\
 Median & 4.33   & 4.33    & 4.33  & 4.33   \\
 Max    & 7.00   & 7.00    & 7.00  & 7.00   \\
 Mean   & 4.26   & 4.25    & 4.26  & 4.25   \\
 std    & 1.95   & 1.94    & 1.95  & 1.95   \\
\end{tabular} \\
\bottomrule
    \end{tabular}
    \caption{Measures of central tendency of the rating values in the regression tasks, overall and per experimental split.}
    \label{tab:labels_4}
\end{table*}

\begin{table*}[ht]
    \centering
    \small
    \begin{tabular}{c|c}
\toprule
     \begin{tabular}{lrrrr}
\multicolumn{5}{c}{\textbf{EmpDial EI.}} \\
               &    All &   Train &   Val &   Test \\
\midrule
 Split          & 100.0 &   80.0 & 10.0 &  10.0 \\
 questioning   &  32.5 &   32.7 & 31.7 &  32.3 \\
 acknowledging &  20.6 &   20.5 & 20.9 &  20.8 \\
 neutral       &  14.7 &   14.8 & 14.4 &  14.4 \\
 agreeing      &  13.0 &   13.0 & 12.9 &  13.2 \\
 sympathizing  &   4.6 &    4.6 &  4.8 &   4.4 \\
 encouraging   &   4.1 &    4.0 &  4.5 &   4.1 \\
 suggesting    &   3.9 &    3.9 &  4.0 &   4.2 \\
 consoling     &   3.7 &    3.7 &  3.8 &   3.9 \\
 wishing       &   2.9 &    2.9 &  3.0 &   2.7 \\
\bottomrule
\end{tabular}
    &  \begin{tabular}{lrrrr}
\multicolumn{5}{c}{\textbf{News Emotion}} \\

          &    All &   Train &   Val &   Test \\
\midrule
 Split     & 100.0 &   29.6 & 37.0 &  33.3 \\
 Sadness  &  38.3 &   39.1 & 37.5 &  38.4 \\
 Neutral  &  27.2 &   26.8 & 27.3 &  27.4 \\
 Anger    &  16.3 &   16.8 & 16.2 &  16.0 \\
 Disgust  &   8.4 &    7.9 &  8.8 &   8.4 \\
 Hope     &   4.2 &    4.0 &  4.4 &   4.1 \\
 Fear     &   3.8 &    3.8 &  3.8 &   3.8 \\
 Joy      &   1.1 &    1.1 &  1.1 &   1.1 \\
 Surprise &   0.7 &    0.6 &  0.9 &   0.7 \\
 &&&&\\
\end{tabular} \\
     \begin{tabular}{lrrrr}
\multicolumn{5}{c}{\textbf{EmpDial Role}} \\
      &    All &   Train &   Val &   Test \\
\midrule
 Split & 100.0 &   80.1 &  9.9 &  10.0 \\
 Speaker   &  62.2 &   62.2 & 62.5 &  62.2 \\
 Listener    &  37.8 &   37.8 & 37.5 &  37.8 \\
\bottomrule
\end{tabular}    &  \begin{tabular}{lrrrr}
\toprule
\multicolumn{5}{c}{\textbf{MI Adherent}} \\
      &    All &   Train &   Val &   Test \\
\midrule
 Split & 100.0 &   79.7 & 10.2 &  10.2 \\
 Adherent    &  68.5 &   68.7 & 67.9 &  67.5 \\
 Non-adherent    &  31.5 &   31.3 & 32.1 &  32.5 \\
\bottomrule
\end{tabular} \\
     \begin{tabular}{lrrrr}
\multicolumn{5}{c}{\textbf{Epitome ER}} \\
      &    All &   Train &   Val &   Test \\
\midrule
 Split & 100.0 &   80.0 & 10.0 &  10.0 \\
 0    &  66.4 &   66.3 & 67.0 &  66.6 \\
 1    &  28.6 &   28.8 & 27.7 &  28.5 \\
 2    &   5.0 &    4.9 &  5.3 &   5.0 \\
\bottomrule
\end{tabular}    &  \begin{tabular}{lrrrr}
\multicolumn{5}{c}{\textbf{Epitome EX}} \\
      &    All &   Train &   Val &   Test \\
\midrule
 Split & 100.0 &   80.0 & 10.0 &  10.0 \\
 0    &  84.4 &   84.4 & 84.5 &  84.1 \\
 1    &   3.4 &    3.4 &  3.3 &   3.3 \\
 2    &  12.2 &   12.2 & 12.2 &  12.6 \\
\end{tabular} \\
 \begin{tabular}{lrrrr}
\multicolumn{5}{c}{\textbf{Epitome IP}} \\
      &    All &   Train &   Val &   Test \\
\midrule
 Split & 100.0 &   80.0 & 10.0 &  10.0 \\
 0    &  52.5 &   52.7 & 51.5 &  52.0 \\
 1    &   3.8 &    3.8 &  3.6 &   3.6 \\
 2    &  43.7 &   43.5 & 44.9 &  44.4 \\
\end{tabular}        & \begin{tabular}{lrrrr}
\toprule
\multicolumn{5}{c}{\textbf{Empathy and Hope}} \\
                &    All &   Train &   Val &   Test \\
\midrule
 Split           & 100.0 &   80.0 & 10.1 &  10.0 \\
 Supportive     &  50.7 &   50.7 & 50.4 &  50.8 \\
 Not Supportive &  49.3 &   49.3 & 49.6 &  49.2 \\
&&&& \\

\end{tabular} \\
\bottomrule
    \end{tabular}
    \caption{Label distributions overall and per each experimental split for the EmpDial EI, EmpDial Role, News Emotion, MI Adherent, Epitome, and Empathy and Hope tasks.}
    \label{tab:labels_3}
\end{table*}

\begin{table*}[ht]
    \centering
    \small
    \begin{tabular}{c|c}
\toprule
      \begin{tabular}{lrrrr}
\multicolumn{5}{c}{\textbf{EmpDial QAct}} \\

                       &    All &   Train &   Val &   Test \\
\midrule
 Split                  & 100.0 &   80.0 & 10.0 &  10.0 \\
 \multirow{2}{*}{\shortstack[l]{Request\\information}}    &  51.4 &   51.5 & 50.3 &  51.3 \\
 &&&& \\
 \multirow{2}{*}{\shortstack[l]{Ask about\\consequence }} &  17.9 &   18.1 & 17.3 &  17.6 \\
 &&&& \\
 \multirow{2}{*}{\shortstack[l]{Ask about\\antecedent }}  &  11.3 &   11.1 & 12.3 &  12.1 \\
 &&&& \\
 \multirow{2}{*}{\shortstack[l]{Suggest a\\solution}}    &   8.0 &    8.0 &  8.2 &   8.0 \\
 &&&& \\
 \multirow{2}{*}{\shortstack[l]{Ask for\\confirmation }} &   5.2 &    5.2 &  4.9 &   5.1 \\
 &&&& \\
 \multirow{2}{*}{\shortstack[l]{Suggest a\\reason}}      &   4.1 &    4.1 &  4.7 &   3.9 \\
 &&&& \\
 \multirow{2}{*}{\shortstack[l]{Positive\\rhetoric}}     &   1.1 &    1.1 &  1.1 &   1.0 \\
 &&&& \\
 \multirow{2}{*}{\shortstack[l]{Negative\\rhetoric}}     &   0.8 &    0.7 &  1.0 &   0.7 \\
 &&&& \\
 Irony                 &   0.2 &    0.2 &  0.1 &   0.1 \\
\end{tabular}  &
\begin{tabular}{lrrrr}
% \toprule
\multicolumn{5}{c}{\textbf{EmpDial QInt}} \\
                    &    All &   Train &   Val &   Test \\
\midrule
 Split               & 100.0 &   80.0 & 10.0 &  10.0 \\
 \multirow{2}{*}{\shortstack[l]{Express\\interest}}    &  60.2 &   60.4 & 59.2 &  59.8 \\
&&&& \\
 \multirow{2}{*}{\shortstack[l]{Express\\concern}}    &  23.4 &   23.2 & 23.7 &  23.9 \\
 &&&& \\
 Sympathize         &   5.1 &    5.0 &  5.9 &   4.9 \\
 Offer relief       &   4.5 &    4.6 &  4.1 &   4.7 \\
 \multirow{2}{*}{\shortstack[l]{Amplify\\excitement}} &   2.3 &    2.3 &  2.7 &   2.2 \\
 &&&& \\
 Support            &   1.0 &    1.0 &  1.0 &   1.0 \\
 Amplify joy        &   0.9 &    0.8 &  0.9 &   1.0 \\
 Amplify pride      &   0.7 &    0.7 &  0.7 &   0.7 \\
 De-escalate        &   0.7 &    0.7 &  0.7 &   0.6 \\
 \multirow{2}{*}{\shortstack[l]{Moralize\\speaker}}    &   0.6 &    0.6 &  0.5 &   0.5 \\
 &&&& \\
 \multirow{2}{*}{\shortstack[l]{Pass\\judgement}}     &   0.5 &    0.5 &  0.4 &   0.5 \\
 &&&& \\
 Motivate           &   0.2 &    0.2 &  0.1 &   0.1 \\
\end{tabular}         \\ 
\bottomrule
    \end{tabular}
    \caption{Label distributions overall and per each experimental split for the EmpDial QAct and QInt tasks.}
    \label{tab:labels_2}
\end{table*}

\begin{table*}[ht]
    \centering
    \small
    \begin{tabular}{c|c}
    \toprule
     \begin{tabular}{lrrrr}
\multicolumn{5}{c}{\textbf{EmpDial Emotion}} \\
              &    All &   Train &   Val &   Test \\
\midrule
 Split         & 100.0 &   80.1 &  9.9 &  10.0 \\
 surprised    &   5.2 &    5.1 &  5.0 &   5.7 \\
 excited      &   3.8 &    3.8 &  3.3 &   3.6 \\
 annoyed      &   3.6 &    3.6 &  3.4 &   3.7 \\
 proud        &   3.5 &    3.5 &  3.1 &   4.1 \\
 angry        &   3.5 &    3.5 &  2.9 &   3.9 \\
 sad          &   3.5 &    3.4 &  3.9 &   3.4 \\
 lonely       &   3.3 &    3.2 &  3.5 &   3.2 \\
 grateful     &   3.2 &    3.2 &  3.4 &   3.2 \\
 afraid       &   3.2 &    3.2 &  3.1 &   3.4 \\
 confident    &   3.2 &    3.2 &  3.2 &   3.4 \\
 disgusted    &   3.2 &    3.2 &  3.2 &   3.2 \\
 impressed    &   3.2 &    3.2 &  2.8 &   3.2 \\
 terrified    &   3.2 &    3.2 &  3.2 &   2.6 \\
 anxious      &   3.1 &    3.2 &  3.0 &   2.9 \\
 disappointed &   3.1 &    3.1 &  3.5 &   3.0 \\
 anticipating &   3.1 &    3.0 &  3.2 &   3.9 \\
 hopeful      &   3.1 &    3.1 &  3.4 &   3.0 \\
 jealous      &   3.1 &    3.1 &  3.4 &   2.8 \\
 guilty       &   3.1 &    3.0 &  3.3 &   3.3 \\
 joyful       &   3.1 &    3.1 &  3.1 &   3.0 \\
 furious      &   3.0 &    3.0 &  2.8 &   3.4 \\
 nostalgic    &   3.0 &    3.0 &  3.2 &   2.9 \\
 prepared     &   3.0 &    3.1 &  2.9 &   2.9 \\
 embarrassed  &   3.0 &    3.0 &  2.8 &   2.7 \\
 content      &   2.9 &    3.0 &  2.9 &   2.5 \\
 sentimental  &   2.8 &    2.9 &  3.0 &   2.6 \\
 devastated   &   2.8 &    2.9 &  2.5 &   2.7 \\
 caring       &   2.7 &    2.7 &  3.0 &   2.9 \\
 trusting     &   2.6 &    2.7 &  2.3 &   2.7 \\
 ashamed      &   2.6 &    2.5 &  3.3 &   2.3 \\
 apprehensive &   2.5 &    2.5 &  2.8 &   2.2 \\
 faithful     &   1.9 &    1.9 &  1.7 &   1.8 \\
\end{tabular}
    &  \begin{tabular}{lrrrr}
\multicolumn{5}{c}{\textbf{MI Behavior}} \\
                           &    All &   Train &   Val &   Test \\
\midrule
 Split                      & 100.0 &   79.7 & 10.2 &  10.2 \\
 \multirow{2}{*}{\shortstack[l]{Give\\Information}}          &  29.9 &   30.4 & 28.0 &  27.9 \\
 &&&& \\
\multirow{2}{*}{\shortstack[l]{Advise w/o\\ Permission}} &  14.3 &   14.0 & 16.0 &  14.9 \\
 &&&& \\
 Self-Disclose             &   8.8 &    8.8 &  7.9 &   9.4 \\
 \multirow{2}{*}{\shortstack[l]{Complex\\Reflection}}        &   8.0 &    7.8 &  9.6 &   8.3 \\
 &&&& \\
 Support                   &   7.8 &    7.9 &  7.5 &   7.7 \\
 Affirm                    &   5.9 &    5.9 &  5.1 &   6.7 \\
 \multirow{2}{*}{\shortstack[l]{Closed\\Question}}           &   5.8 &    5.8 &  5.7 &   5.3 \\
 &&&& \\
 Direct                    &   5.7 &    5.8 &  5.0 &   5.7 \\
 \multirow{2}{*}{\shortstack[l]{Simple\\Reflection}}       &   3.5 &    3.4 &  4.1 &   3.4 \\
 &&&& \\
 Open Question             &   3.0 &    3.0 &  3.3 &   3.2 \\
  \multirow{2}{*}{\shortstack[l]{Advise w/\\Permission}}     &   3.0 &    2.9 &  3.3 &   3.4 \\
  &&&& \\
 Confront                  &   2.0 &    2.0 &  2.4 &   1.8 \\
 \multirow{2}{*}{\shortstack[l]{Emphasize\\Autonomy}}    &   1.6 &    1.6 &  1.4 &   1.6 \\
 &&&& \\
 Warn                      &   0.7 &    0.7 &  0.8 &   0.8 \\
 &&&& \\
 &&&& \\
 &&&& \\
 &&&& \\
 &&&& \\
 &&&& \\
 &&&& \\
 &&&& \\
 &&&& \\
 &&&& \\
 &&&& \\
\end{tabular} \\
\bottomrule
    \end{tabular}
    \caption{Label distributions overall and per each experimental split for the EmpDial Emotion and MI Behavior tasks.}
    \label{tab:labels_1}
\end{table*}